\documentclass[acmtog]{acmart}

\usepackage{microtype}
\usepackage{subcaption}
\usepackage{appendix}

\hypersetup{
  pageanchor=true,
  plainpages=false,
  pdfpagelabels=true
}

\AtEndDocument{\label{TotPages}}

\setcopyright{rightsretained}
\copyrightyear{2025}
\acmYear{2025}
\acmDOI{https://rexera.github.io}
 
\acmJournal{FACMP}
\acmVolume{ }
\acmNumber{ }
\acmArticle{ }

\acmSubmissionID{123-A56-BU3}
 
\begin{document}

\title{Contrastive Analysis of Constituent Order Preferences Within Adverbial Roles in English and Chinese News: A Large-Language-Model-Driven Approach}

\author{Yiran Rex Ma}
\email{mayiran@bupt.edu.cn}
\orcid{0009-0002-3440-9819}
\affiliation{%
  \institution{School of Humanities, Beijing University of Posts and Telecommunications}
  \state{Beijing}
  \country{China}
}

\begin{abstract}
  Based on comparable English-Chinese news corpora annotated by Large Language Model (LLM), this paper attempts to explore the differences in constituent order of English-Chinese news from the perspective of functional chunks with adverbial roles, and analyze their typical positional preferences and distribution patterns. It is found that: (1) English news prefers linear narrative of “core information first”, and functional chunks are mostly post-positioned, while Chinese news prefers overall presentation mode of “background first”, and functional chunks are often pre-positioned; (2) In SVO structure, both English and Chinese news show differences in the distribution of functional chunks, but the tendency of Chinese pre-positioning is more significant, while that of English post-positioning is relatively mild; (3) When function blocks are co-occurring, both English and Chinese news show high flexibility, and the order adjustment is driven by information and pragmatic purposes. The study reveals that word order has both systematic preference and dynamic adaptability, providing new empirical support for contrastive study of English-Chinese information structure.\footnote{Project repository available at: \url{https://github.com/rexera/constituent_order}}
\end{abstract}




\maketitle

\section{Introduction}

Constituent order refers to the sequence of words or phrases as functional units within a syntactic structure \cite{jacobson1996,anyuxia2006}. It is a subset of word order. Differences in constituent order are evident in English and Chinese sentences, particularly when conveying adverbial information such as time, location, and manner. These differences pose challenges for language learning, communication, and practical use. The primary difficulties include the following: (1) Second-language learners often struggle to accurately determine constituent placement \cite{lixijiang2020,huangyiqing2010}. For example, as illustrated in examples (1)a and (1)b, there are significant word order differences when delivering the same news information, showing clear preferences. (2) Translators need to adjust word order to maintain effective information transmission \cite{zhanglu2019}. For instance, as demonstrated in examples (1)b and (1)d, translators must align with the target audience’s expectations regarding word order. While existing studies provide qualitative assessments of constituent order preferences \cite{yuanyulin1999,wuweizhang1995}, they fall short of offering systematic patterns of constituent usage.

\begin{figure}[htbp]
  \centering
  \includegraphics[width=0.5\textwidth]{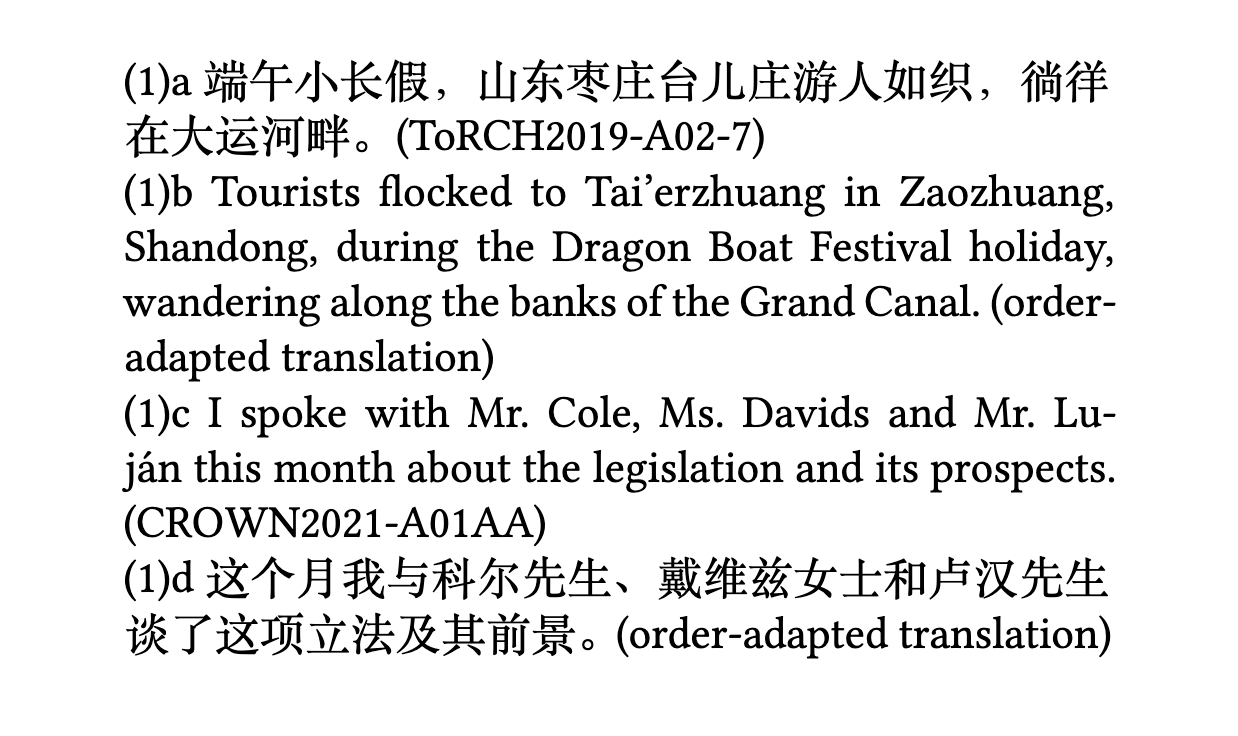}
\label{fig:quote1}
\end{figure}

Scholars have engaged in extensive discussions on word order and the sequencing of adverbials, focusing primarily on the following areas: (1) the manifestation of word order in grammatical units such as attributes, adverbials, and subordinate clauses \cite{huyushu1988}; (2) grammatical and cognitive factors influencing word order and adverbial placement \cite{wasow2003}; and (3) cross-linguistic similarities and differences in word order and whether universal patterns exist \cite{hawkins1994,rongjing2000}.

Previous research exhibits the following characteristics:

First, in terms of analytical focus, studies often begin with specific grammatical units, emphasizing syntactic analysis but paying insufficient attention to semantic information transmission \cite{lijinman2010,zengjinghan2015,muller2021}. Second, regarding methodology, many investigations rely on inductive reasoning and subjective insights to infer word order differences, often constrained by limited data and lacking empirical support for universal patterns \cite{guozhong2012,shuangwenting2016,zhang2022}. Third, in research priorities, while some studies highlight factors influencing word order differences, less attention is given to the distributional patterns of word order itself \cite{futrell2020}.

\citet{huyushu1988} define constituent order as the relationship and arrangement of  ``functional chunks.'' Different languages, when expressing the same semantic function, may make similar or divergent choices in arranging these blocks based on their syntactic and semantic rules. With advancements in large language models such as GPT-4 \cite{openai2023}, the automated annotation of ``functional chunks'' within corpora has enabled the efficient analysis of large-scale word order patterns. Building on this, the present study employs a data-driven approach using automated annotation methods provided by large language models. It focuses on the distributional patterns of adverbial functional chunks in English and Chinese journalistic texts, aiming to uncover the preferences for constituent order in both languages.

\section{Research Background}
 
\subsection{Typological Representation of Word Order Differences in English and Chinese}
Human languages universally exhibit correlations between syntactic positions and word order typology \cite{dryer1986}. For instance, although both English and Chinese share an SVO structure\footnote{English sentence structures are typically categorized into seven patterns: SV, SVO, SVC, SVA, SVOO, SVOC, and SVOA \cite{quirk1985}, with the SVO pattern being the most common. For Chinese, interpretations differ. One view asserts that subject-verb-object (SVO) is the basic word order of Modern Chinese \cite{wangli1980}. Another perspective argues that as a character-based language, Chinese cannot be fully represented by the SVO framework and is better described using a "theme-rheme" structure \cite{zhaoyuanren1968,zhangbojiang2011}. This study does not delve further into this debate. For the sake of comparison, we provi-sionally assume that both English and Chinese share SVO as the most prevalent sentence pattern.}, they differ in the placement of modifiers: English tends to position modifiers after the headword, while Chinese generally places them before it \cite{wangkangmao1981}. When conveying temporal, locative, and manner information, English and Chinese exhibit distinct preferences for constituent placement within sentences. These differences are prevalent across various grammatical units, including attributes, adverbials, and subordinate clauses \cite{anyuxia2006}.

Existing scholarly discussions often provide a coarse-grained categorization of these differences, offering limited insights into a comprehensive understanding of word order variations. While there is consensus that English and Chinese differ typologically in terms of word order, intuitive typological generalizations still dominate discussions, particularly regarding the flexible placement of adverbials. This makes it difficult to extract typical distributional patterns of word order \cite{dingzhibin2014}.

Regarding the relationship between word order and syntax, \citet{lijinman2010} observed that although Chinese is primarily SVO, it demonstrates an OV tendency in relative clauses and comparative structures, resulting in a typologically atypical mixed-order feature among global languages. Using word order indices, \citet{jinlixin2019} further confirmed that while Chinese exhibits VO characteristics in verb-object structures, it displays OV tendencies in areas such as attributive and prepositional phrase placement. Additionally, Chinese relies on a ``topic chain'' to advance information, allowing for layered information expansion and permitting adverbials to appear at the beginning of sentences. In contrast, English more rigidly adheres to the subject-verb-object logical order \cite{rongjing2000}.

However, these conclusions are predominantly based on introspective reasoning and speculative analysis, with insufficient empirical validation through large-scale corpora.

\subsection{Semantic and Information-Structural Motivations}
In the evolution of human languages, a universal tendency toward ``unidirectional change'' aims to enhance the efficiency of syntactic processing \cite{hawkins1994}. This shared tendency leads to certain commonalities between English and Chinese in discourse-level information progression. For instance, \citet{liu2014} and \citet{guohua2022}, applying \citet{halliday1985}'s theme and rheme theory to broadcast and academic discourse, respectively, identified several shared thematic progression patterns in both languages.

However, English and Chinese have diverged over time. English often adopts a front-loaded structure where the focus is placed early in the sentence, while Chinese prefers an end-weighted structure that reveals the core information toward the sentence's conclusion \cite{jiadelin1990}. English word order is constrained by syntactic structures, movement operations, and principles such as linear modification, with information typically organized in a given-to-new or weak-to-strong sequence \cite{chamonikolasova2009,lee2019}. Cognitive principles like dependency locality and extraposition are also employed to reduce syntactic complexity and cognitive load \cite{huddleston2002,futrell2020}. For example, \citet{hannay1991} proposed information management strategies, including frameworks like all-new pattern, topic pattern, and reaction pattern, emphasizing their ties to constituent frequency and cognitive entrenchment.

In contrast, Chinese word order plays a greater grammatical role due to the lack of morphological inflections and is influenced by multiple factors, including syntactic structures, information structure, and cognitive iconicity \cite{anyuxia2006}. Chinese exhibits traits such as end-weighted information focus, frequent omission of given information, and a old-to-new information sequence \cite{zhanglianqiang1997,rongjing2000}.

The flexibility of Chinese word order remains a subject of scholarly debate. Certain syntactic constraint and valency studies argue that the ``word order-function'' relationship is relatively fixed, with information placement governed by the flow of information, scope of modifiers, or control of semantic roles \cite{ross1984,xing2015}. However, other researchers highlight the dynamic nature of word order. For example, subordinate structures in Chinese often follow a head-final pattern, driven by pragmatic demands and semantic compatibility \cite{yurayong2023}.

Given these complexities, there is a need to analyze the semantic and information-structural motivations behind word order differences in journalistic texts. By leveraging large-scale comparable corpora, this study aims to explore the distinct constituent order preferences in English and Chinese news discourse.

\subsection{The Trend of Patterned Word Order Distribution}
Previous studies on English word order have highlighted that the placement of adverbials is not random but is often regulated by principles of information structure, leading to positional preferences \cite{stolterfoht2013}. Similarly, \citet{biber1999} observed that different types of adverbials exhibit specific distributional tendencies within sentences, and these tendencies are significantly influenced by variations in registers. For Chinese, \citet{zhang2022} analyzed the positional distribution of temporal adverbial clauses using dependency treebanks, examining their syntactic and semantic characteristics. Collectively, these studies reveal a trend toward patterned constituent order distribution in both English and Chinese corpora.

Traditional corpus-based research has largely relied on manual annotation \cite{ai2016}. Due to limitations in data volume, these studies often struggle to extract broader generalizations regarding word order patterns. For instance, \citet{travis1984}, using the Domain Adjacency Condition in parametric syntax theory, explained constraints on the arrangement of verbs and adverbials but lacked empirical validation. Similarly, the Dependency Locality model proposed by \citet{futrell2020} predicts word order based on dependency length but falls short in explicitly extracting distributional rules.

The advent of large language models has provided new opportunities for advancing research in this area. For example, \citet{kojima2022} introduced the zero-shot chain-of-thought (0-Shot CoT) approach, which demonstrates the potential of large models for automated annotation of large-scale corpora. This innovation offers a powerful tool for exploratory data analysis of constituent order differences.

To comprehensively uncover the distributional patterns of constituent order in English and Chinese, this study employs a comparable corpus of journalistic texts. Leveraging the automatic annotation capabilities of advanced language models, it aims to systematically analyze and compare constituent order preferences across the two languages.

\section{Research Design}
\subsection{Research Questions}
Focusing on journalistic texts, this study investigates the intralingual preferences and interlingual differences in English and Chinese constituent order from the perspective of adverbial functional chunks. Specifically, it addresses the following questions:

(1) What are the word order preferences for different types of functional chunks?

(2) What are the trends in constituent order within SVO-functional chunk combinations?

(3) What are the sequential tendencies when multiple functional chunks co-occur?

\subsection{Corpus Sources and Annotation}
This study utilizes the CROWN2021 \cite{sun2022} and ToRCH2019 \cite{lijialei2022} sub-corpora A (hereinafter referred to as CROWN and ToRCH), which are well-represented collections of journalistic texts. Both corpora are built on the BROWN family corpus paradigm, ensuring cross-linguistic comparability. 

CROWN2021 is a balanced corpus of American English compiled by Beijing Foreign Studies University in 2021, while ToRCH2019 is a balanced corpus of Modern Mandarin Chinese compiled in 2019. Each corpus comprises approximately 1 million words. For this study, the English subcorpus contains 90,131 words, and the Chinese sub-corpus includes 88,539 words. Relevant statistics are shown in Table \ref{tab:table3}.

Based on the semantic functions of adverbials, the study follows the classification proposed by \citet{quirk1985} and categorizes functional chunks (FCs) into eight types: <time>, <place>, <manner>, <cause>, <effect>, <condition>, <purpose>, and <concession>. Additionally, subject, verb, and object are annotated as <S>, <V>, and <O> to ensure consistency with the SVO structure in both English and Chinese. For subordinate clauses, each is treated as a separate SVO structure. Words or phrases that cannot be classified into these categories are left unannotated, allowing for gaps in annotation.

To achieve a balance between efficiency and reliability, this study employs the GPT-4o model API\footnote{https://api.openai.com/v1/fine\_tuning/jobs; Model: \texttt{gpt-4o-2024-08-06}}, recognized for its superior performance, and incorporates a few-shot prompting strategy to guide the model's annotation according to the predefined labeling scheme\footnote{The effectiveness and limitations of automatic annotation using large language models have been discussed in academic research. While these models can achieve, and sometimes surpass, the capabilities of crowdsourced annotators \cite{alizadeh2024} and offer strong scalability, they still exhibit issues such as bias, hallucinations, and limited expertise in specialized domains. Methods such as enhancing data diversity and incorporating model self-correction mechanisms have been shown to mitigate these challenges \cite{tan2024}. In response, this study adopts a few-shot prompting approach, incorporating three examples comparing original sentences with manually annotated results. The prompt also includes examples for each type of functional block to clarify distinc-tions. Although the method does not achieve human-expert-level quality for sentence-by-sentence annotation, it ensures usability for macro-level statistical analysis. The annotation process was conducted in parallel twice, simulating a dual-annotator strategy. Independent annotation of the same dataset was followed by a consistency check, which yielded an agreement rate exceeding 0.95.}. Examples of annotated sentences and their abstracted versions are shown in Example (2), while annotation statistics are presented in Tables \ref{tab:table4}, \ref{tab:table5}, \ref{tab:table6}.

\begin{figure}[h]
  \centering
  \includegraphics[width=0.5\textwidth]{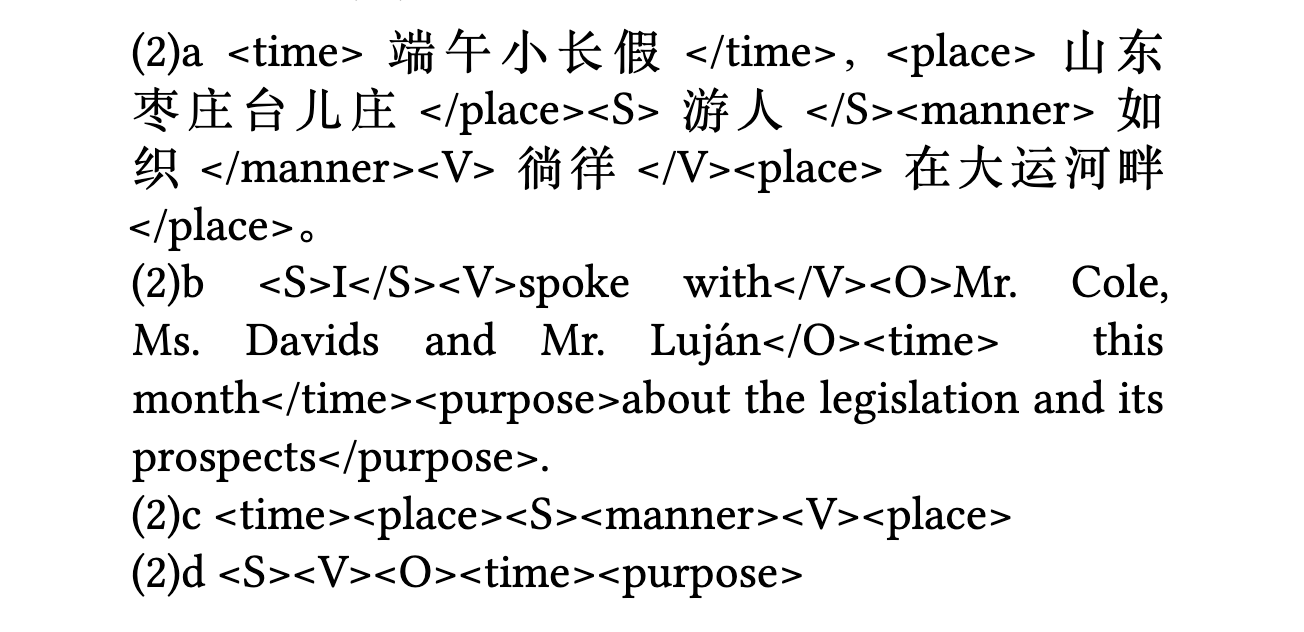}
  \label{fig:quote2}
\end{figure}

\begin{table}[htbp]
\centering
\caption{Statistics of Original Data}
\label{tab:table3}
\begin{tabular}{lllll}
\toprule
Corpus & Texts & Tokens & Types & TTR \\
\midrule
CROWN & 220 & 90131 & 12502 & 0.139 \\
ToRCH & 44 & 88539 & 12341 & 0.139 \\
\bottomrule
\end{tabular}
\end{table}

\begin{table}[htbp]
\centering
\caption{Statistics of Annotated Data}
\label{tab:table4}
\begin{tabular}{llllll}
\toprule
Corpus & Lines & Tags & Tag/Line & FCs & FC/Line \\
\midrule
CROWN & 2649 & 17865 & 6.74 & 5846 & 2.21 \\
ToRCH & 1735 & 26162 & 15.08 & 8389 & 4.84 \\
\bottomrule
\end{tabular}
\end{table}

\begin{table}[htbp]
\centering
\caption{Distribution of Functional Chunks in English News Discourse}
\label{tab:table5}
\begin{tabular}{lll}
\toprule
Tag & Frequency & Proportion \\
\midrule
<time> & 1326 & 0.23 \\
<place> & 1129 & 0.19 \\
<effect> & 1098 & 0.19 \\
<purpose> & 765 & 0.13 \\
<manner> & 540 & 0.09 \\
<cause> & 476 & 0.08 \\
<concession> & 356 & 0.06 \\
<condition> & 156 & 0.03 \\
\bottomrule
\end{tabular}
\end{table}

\begin{table}[htbp]
\centering
\caption{Distribution of Functional Chunks in Chinese News Discourse}
\label{tab:table6}
\begin{tabular}{lll}
\toprule
Tag & Frequency & Proportion \\
\midrule
<manner> & 3293 & 0.39 \\
<place> & 2362 & 0.28 \\
<time> & 2050 & 0.24 \\
<effect> & 235 & 0.03 \\
<concession> & 168 & 0.02 \\
<purpose> & 112 & 0.01 \\
<condition> & 93 & 0.01 \\
<cause> & 76 & 0.01 \\
\bottomrule
\end{tabular}
\end{table}
\subsection{Statistical Analysis}
This study employs statistical methods to investigate the distributional characteristics and word order patterns of functional chunks in English and Chinese journalistic texts, aiming to uncover differences in constituent order preferences between the two languages. According to construction grammar \cite{goldberg1999}, constructions and their internal components exhibit cognitive entrenchment, which arises from categorization and pattern formation through high-frequency language input. Similarly, the close relationship between constituent order and information structure can be seen as a form of entrenched and generalized preference.

In this process, constituent order gradually adapts to its semantic functions. At the statistical level, the preferences for constituent order deviate significantly from the assumption of uniform distribution when achieving the same informational functions. From this perspective, if the relative position of functional chunks within a sentence shows significant differences ($p$ < 0.05) from the uniform distribution assumption both intralingually and interlingually, it indicates that positional preferences are systematic and stable within a language and exhibit notable variation across languages.

To explore these phenomena, this study combines conditional probability analysis, chi-square tests, independent sample t-tests, and Markov chain transition matrix analysis to examine the following aspects:

(1) Sentence-internal positional preferences of different types of functional chunks;

(2) Combination patterns of functional chunks within the SVO framework;

(3) Sequential patterns and distributional tendencies of multiple functional chunks.

\section{Research Findings}
\subsection{Sentence-Internal Distribution of Different Types of Functional Chunks}
\begin{figure*}[htbp]
\centering
\caption{Relative Position Distribution of Various Functional Chunks in Chinese Journalistic Texts}
\vspace{0.3cm}
\begin{subfigure}[b]{0.24\linewidth}
\includegraphics[width=\linewidth]{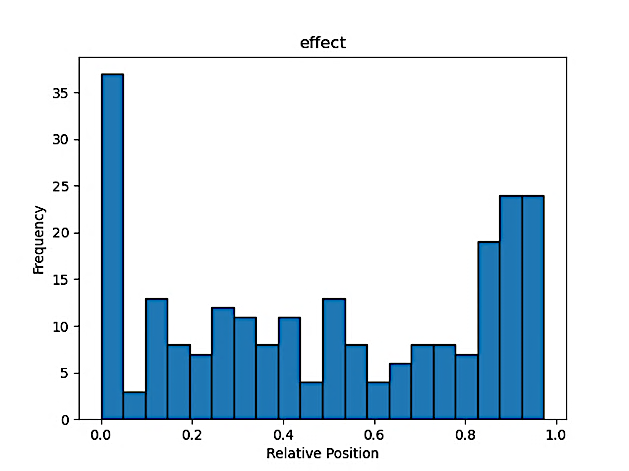}
\label{fig:sub1}
\end{subfigure}
\hfill
\begin{subfigure}[b]{0.24\linewidth}
\includegraphics[width=\linewidth]{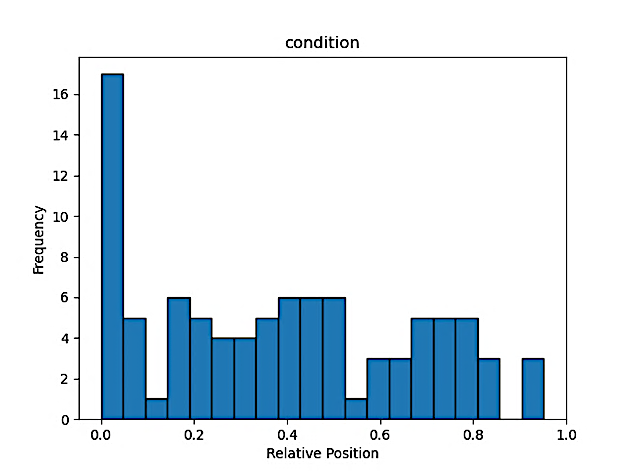}
\label{fig:sub2}
\end{subfigure}
\hfill
\begin{subfigure}[b]{0.24\linewidth}
\includegraphics[width=\linewidth]{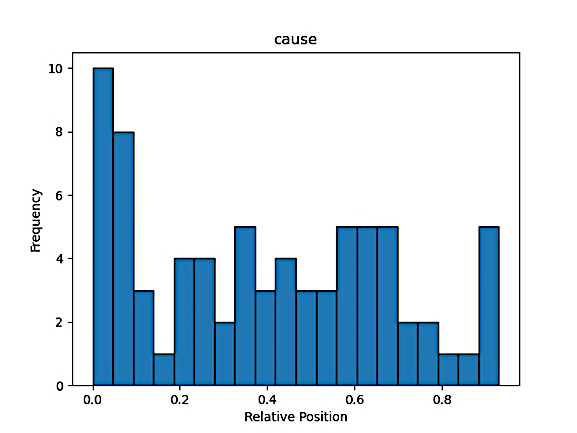}
\label{fig:sub3}
\end{subfigure}
\hfill
\begin{subfigure}[b]{0.24\linewidth}
\includegraphics[width=\linewidth]{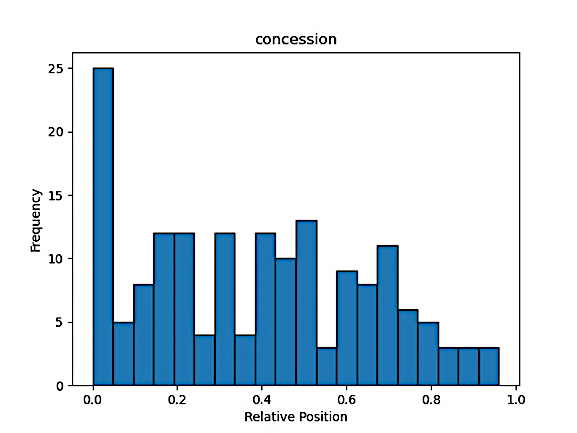}
\label{fig:sub4}
\end{subfigure}

\vspace{0.3cm}

\begin{subfigure}[b]{0.24\linewidth}
\includegraphics[width=\linewidth]{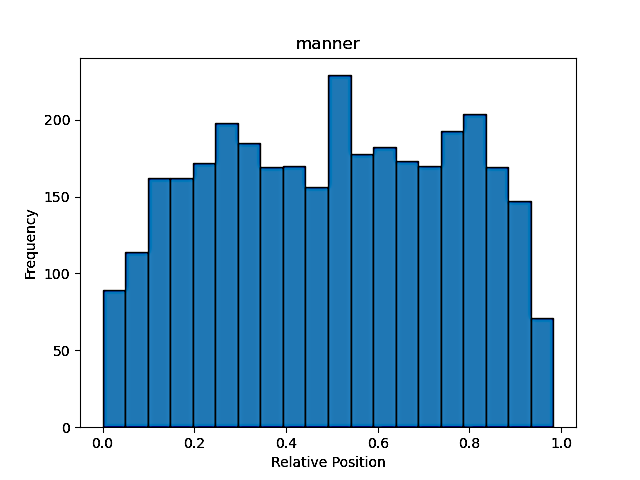}
\label{fig:sub5}
\end{subfigure}
\hfill
\begin{subfigure}[b]{0.24\linewidth}
\includegraphics[width=\linewidth]{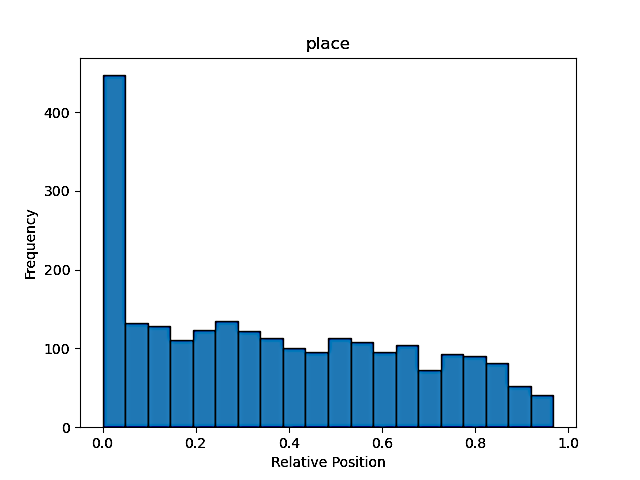}
\label{fig:sub6}
\end{subfigure}
\hfill
\begin{subfigure}[b]{0.24\linewidth}
\includegraphics[width=\linewidth]{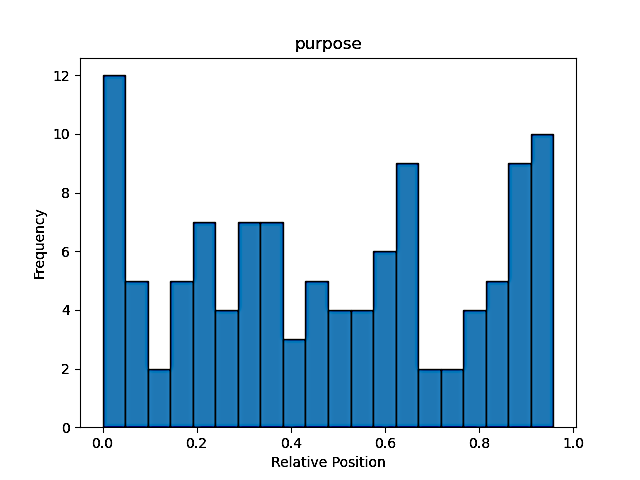}
\label{fig:sub7}
\end{subfigure}
\hfill
\begin{subfigure}[b]{0.24\linewidth}
\includegraphics[width=\linewidth]{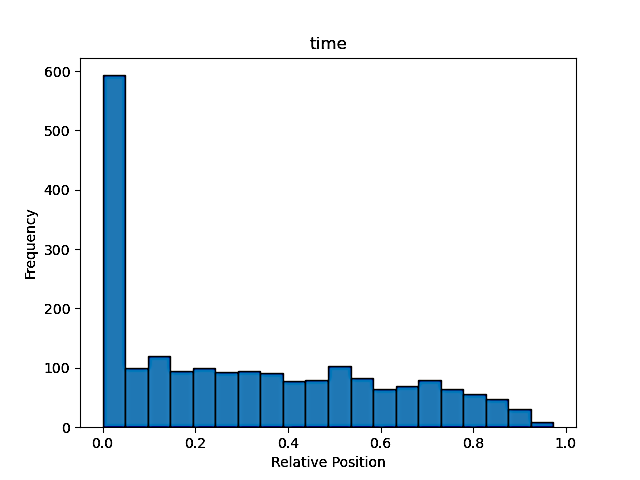}
\label{fig:sub8}
\end{subfigure}

\label{fig:combined_images}
\end{figure*}

First, chi-square tests were conducted for each type of functional chunk within each language to compare their distribution against the assumption of uniform distribution (see Table \ref{tab:table7}). In Chinese journalistic texts, functional chunks exhibit an overall significant tendency for fronting within sentences (see Figure \ref{fig:combined_images}). Among the chi-square test results for individual functional chunks, <time> and <place> are highly concentrated at the beginning of sentences, often establishing temporal and spatial contexts in advance for the narrative. Additionally, <condition>, <concession>, and <cause> chunks are also primarily positioned at the sentence-initial position, providing logical scaffolding for sentence development.

In contrast, other functional chunks, such as <manner>, <effect>, and <purpose>, display relatively less significant positional preferences. As shown in the bar chart, their sentence-internal distributions do not exhibit strong tendencies. These distributional preferences highlight specific characteristics of information organization in Chinese journalistic discourse. Highly fronted chunks like <time>, <place>, <condition>, and <cause> contribute to constructing a clear contextual framework for narration. Meanwhile, other functional chunks are distributed more flexibly, adapting to various contextual needs. This pattern aligns closely with the emphasis on holistic information structuring and preparatory elaboration commonly observed in Chinese.

\begin{figure*}[htbp]
\centering
\caption{Relative Position Distribution of Various Functional Chunks in English Journalistic Texts}
\vspace{0.3cm}
\begin{subfigure}[b]{0.24\linewidth}
\includegraphics[width=\linewidth]{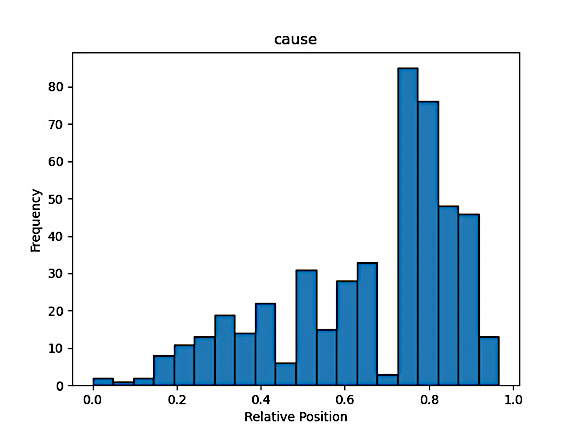}
\label{fig:sub9}
\end{subfigure}
\hfill
\begin{subfigure}[b]{0.24\linewidth}
\includegraphics[width=\linewidth]{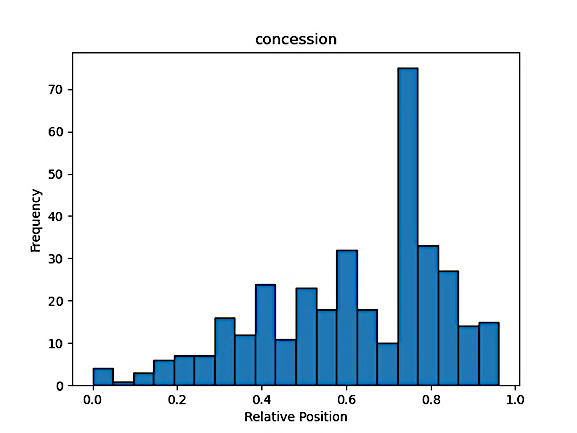}
\label{fig:sub10}
\end{subfigure}
\hfill
\begin{subfigure}[b]{0.24\linewidth}
\includegraphics[width=\linewidth]{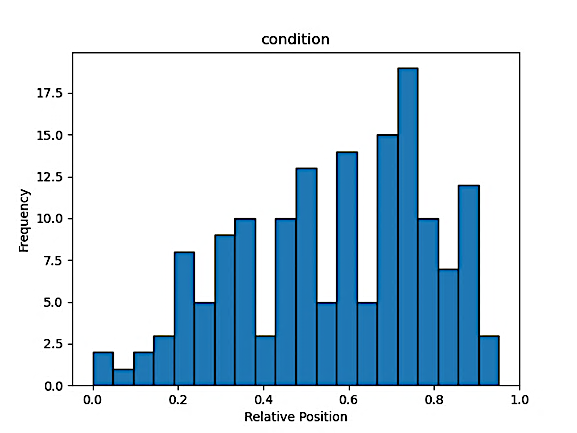}
\label{fig:sub11}
\end{subfigure}
\hfill
\begin{subfigure}[b]{0.24\linewidth}
\includegraphics[width=\linewidth]{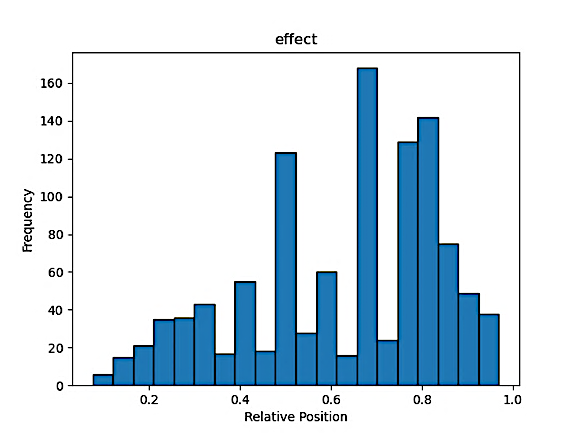}
\label{fig:sub12}
\end{subfigure}

\vspace{0.3cm}

\begin{subfigure}[b]{0.24\linewidth}
\includegraphics[width=\linewidth]{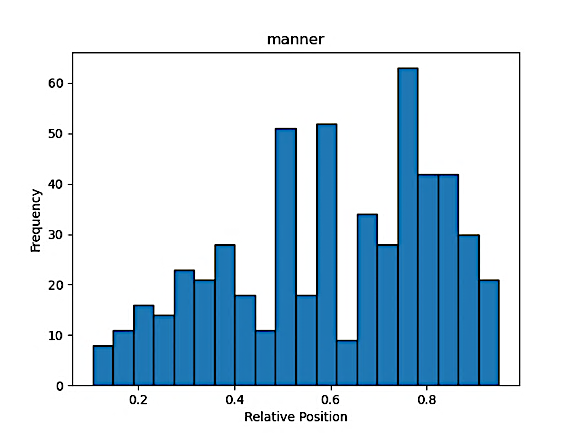}
\label{fig:sub13}
\end{subfigure}
\hfill
\begin{subfigure}[b]{0.24\linewidth}
\includegraphics[width=\linewidth]{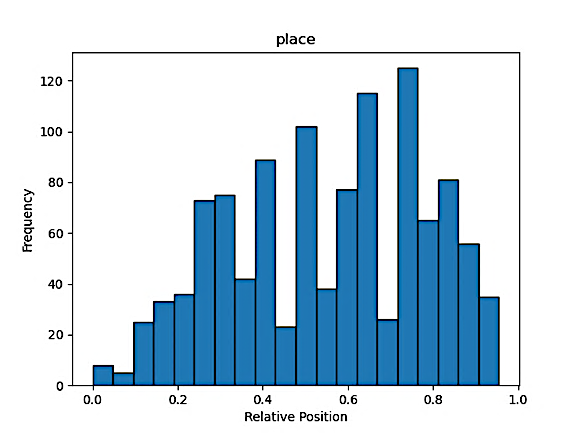}
\label{fig:sub14}
\end{subfigure}
\hfill
\begin{subfigure}[b]{0.24\linewidth}
\includegraphics[width=\linewidth]{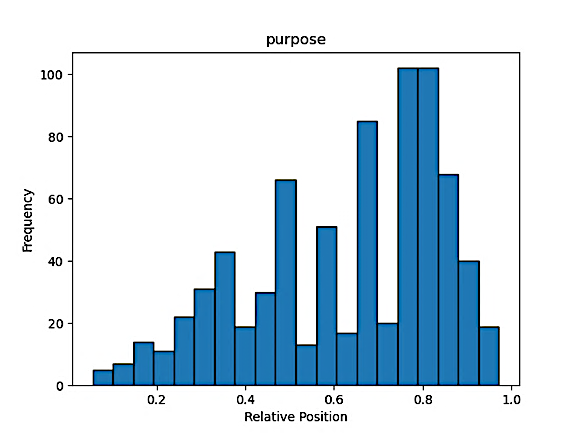}
\label{fig:sub15}
\end{subfigure}
\hfill
\begin{subfigure}[b]{0.24\linewidth}
\includegraphics[width=\linewidth]{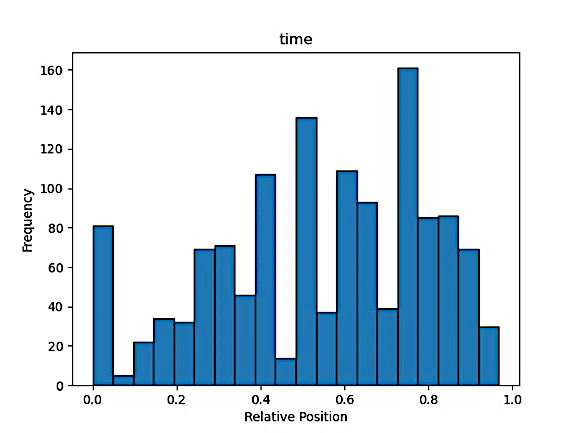}
\label{fig:sub16}
\end{subfigure}

\label{fig:combined_images_9_16}
\end{figure*}

In English journalistic texts, the sentence-internal distribution of functional chunks also shows significant preferences, with an overall tendency toward post-positioning, which contrasts sharply with Chinese (see Figure \ref{fig:combined_images_9_16}). Functional chunks such as <effect>, <purpose>, <cause>, and <concession>, which emphasize conditional or logical relationships, are typically placed at the end of sentences, highlighting their role in providing explanatory support for the core information.

Similarly, <time> and <place> chunks are also significantly concentrated at the sentence-final position, reflecting the English preference for linear progression of temporal and spatial information. In contrast, <condition> exhibits weaker distributional variation but still demonstrates a statistically significant tendency toward sentence-medial or sentence-final positioning.

These distributional patterns illustrate the distinct characteristics of information organization in English journalistic texts. Functional chunks are positioned later in the sentence to make room for the core focal information at the beginning, while maintaining a logical linear narrative flow. This ensures that the key events or information gain greater cognitive prominence at the forefront of the sentence.

In the cross-linguistic comparison, all functional chunks exhibit significant differences under the t-test, with $p$ < 0.05 for each (see Table \ref{tab:table7}). The most notable differences are observed in the distributions of <time> and <place>. In Chinese, these chunks are typically placed at the beginning of sentences to establish temporal and spatial contexts for the narrative, whereas in English, they are more often positioned at the end, aligning with the linear progression of information.

The handling of <manner> information is more flexible in Chinese, while English demonstrates a stronger preference for post-positioning. Functional chunks conveying logical relationships also show significant differences: Chinese tends to emphasize logical prelude, whereas English favors the post-positioning of logical elements for subsequent elaboration.

Overall, the sentence-internal distribution of functional chunks in English and Chinese reveals consistent and systematic preferences and patterns across both languages.

\begin{table}[htbp]
\centering
\caption{Chi-Square Test and Cross-Linguistic t-Test Results for Functional Chunk Distribution}
\label{tab:table7}
\begin{tabular}{llll}
\toprule
FC & Chinese ($\chi^2$,$p$) & English ($\chi^2$,$p$) & C-E($p$) \\
\midrule
<time> & 612.81, <0.001 & 18.99, <0.001 & <0.001 \\
<place> & 332.63, <0.001 & 31.29, <0.001 & <0.001 \\
<condition> & 10.18, 0.001 & 4.14, 0.042 & <0.001 \\
<concession> & 17.32, <0.001 & 40.07, <0.001 & <0.001 \\
<cause> & 5.99, 0.014 & 94.76, <0.001 & <0.001 \\
<manner> & 0.13, 0.72 & 42.59, <0.001 & <0.001 \\
<effect> & 0.018, 0.89 & 135.62, <0.001 & <0.001 \\
<purpose> & 0.28, 0.60 & 100.24, <0.001 & <0.001 \\
\bottomrule
\end{tabular}
\end{table}
\subsection{Comparison of Functional Chunk and SVO Combination Patterns}
To explore the relationship between functional chunks and the SVO framework, we calculated the conditional probability of each type of functional chunk appearing before or after S, V, and O. Taking the example of the conditional probability of <time> appearing before <S>, we treated “the sentence contains a <time> functional chunk” as the conditional event and “<time> appears before <S>” as the target event. This probability represents “the likelihood of <time> being positioned before <S> when serving its intended function.”

The analysis of Chinese data shows that functional chunk distributions are more concentrated, with a clear and consistent preference for fronting in relation to S, V, and O (see Table \ref{tab:table8}). Among the functional chunks with significant differences, <time>, <place>, and <cause> are especially prominent in their distribution before verbs and objects. For chunks with less significant differences, such as <manner> and <condition>, their distribution before and after subjects is more flexible. This flexibility may relate to their role as modifiers within sentences. For instance, <manner> can modify either the subject or the verb, leading to a relatively balanced distribution before and after the subject.

Overall, Chinese functional chunks exhibit a highly unified fronting preference when combined with S, V, and O. However, <manner> and <condition> show greater flexibility around the subject position. These distributional patterns reflect the interplay between a background-first approach and flexible narrative organization in Chinese information structuring.

\begin{table}[htbp]
\centering
\caption{Conditional Probability Table for Chinese Functional Chunk–SVO Distribution}
\label{tab:table8}
\begin{tabular}{lllllll}
\toprule
FC & *<S> & <S>* & *<V> & <V>* & *<O> & <O>* \\
\midrule
<manner> & 0.57 & 0.43 & 0.88 & 0.12 & 0.93 & 0.07 \\
<place> & 0.72 & 0.28 & 0.94 & 0.06 & 0.94 & 0.06 \\
<effect> & 0.58 & 0.42 & 0.67 & 0.33 & 0.72 & 0.28 \\
<time> & 0.81 & 0.19 & 0.98 & 0.03 & 0.97 & 0.03 \\
<purpose> & 0.61 & 0.39 & 0.81 & 0.19 & 0.84 & 0.17 \\
<cause> & 0.78 & 0.22 & 0.96 & 0.04 & 0.92 & 0.08 \\
<condition> & 0.69 & 0.31 & 0.93 & 0.08 & 0.89 & 0.11 \\
<concession> & 0.77 & 0.23 & 0.96 & 0.04 & 0.95 & 0.05 \\
\bottomrule
\end{tabular}
\end{table}
In contrast, the distribution of functional chunks in English exhibits greater flexibility, with smaller differences in conditional probabilities between their positions before and after S, V, and O (see Table \ref{tab:table9}). Functional chunks such as <effect>, <cause>, and <purpose> are slightly more likely to occur after verbs and objects than before. Meanwhile, <time>, <place>, and <manner> demonstrate a more balanced distribution, with minimal differences between their pre- and post-positions. This balance likely reflects the dual function of these chunks: they can modify information at the beginning of a sentence or provide supplementary details about actions or objects at the end, showcasing their flexibility.

Overall, the distribution of functional chunks in English is more even compared to Chinese. Functional chunks such as <effect>, <cause>, and <purpose> exhibit a certain tendency toward post-positioning, while <time>, <place>, and <manner> are distributed more flexibly, with minimal differences between pre- and post-positions. This distribution pattern reflects the diverse placement of functional chunks in English news discourse, offering greater freedom for narrative structuring. It also highlights the emphasis on logical progression and the prominence of syntactic cores in English.

\begin{table}[htbp]
\centering
\caption{Conditional Probability Table for English Functional Chunk–SVO Distribution}
\label{tab:table9}
\begin{tabular}{lllllll}
\toprule
FC & *<S> & <S>* & *<V> & <V>* & *<O> & <O>* \\
\midrule
<manner> & 0.37 & 0.63 & 0.40 & 0.60 & 0.32 & 0.68 \\
<place> & 0.39 & 0.61 & 0.44 & 0.57 & 0.37 & 0.63 \\
<effect> & 0.34 & 0.66 & 0.33 & 0.67 & 0.31 & 0.69 \\
<time> & 0.40 & 0.60 & 0.44 & 0.56 & 0.41 & 0.59 \\
<purpose> & 0.35 & 0.65 & 0.35 & 0.65 & 0.29 & 0.71 \\
<cause> & 0.32 & 0.68 & 0.31 & 0.69 & 0.27 & 0.73 \\
<condition> & 0.45 & 0.55 & 0.47 & 0.53 & 0.38 & 0.62 \\
<concession> & 0.37 & 0.63 & 0.40 & 0.60 & 0.33 & 0.67 \\
\bottomrule
\end{tabular}
\end{table}

Secondly, we extracted two sets of high-frequency functional chunk–SVO distribution patterns from the data (see Appendix Tables \ref{tab:table17} and \ref{tab:table18}). In Chinese texts, <place><S><V><O> is a typical distribution pattern. For instance, in example (3)a, multiple consecutive <place> chunks appear before the subject, establishing a rich and specific spatial background. This arrangement reinforces the ``background-to-focus'' logical structure commonly seen in Chinese narratives.

Another pattern, <time><place><S><V><O>, demonstrates the combined fronting tendency of temporal and spatial functional chunks. In example (3)b, the <time> and <place> chunks consecutively appear at the beginning of the sentence, clarifying the temporal and spatial scope of the narrative. This combination pattern further highlights the regularity of functional chunk arrangements in Chinese news: dual functional chunks strengthen the informational background, providing ample context and depth for the main clause’s narrative.

A third typical pattern is <place><S><manner><V><O>. For example, in (3)c, the <place> chunk at the sentence's start introduces the scene, while the <manner> chunk is inserted between the subject and the verb, modifying the manner of the action. This flexible combination reflects the versatility of functional chunk distribution in Chinese news, consistent with the observation that <manner> chunks can occur both before and after the subject.

These typical patterns demonstrate that the combination of functional chunks and SVO in Chinese not only exhibits an overall fronting preference but also reveals characteristics of joint fronting and flexibility. This partly uncovers the underlying regularities of information backgrounding and narrative logic in Chinese journalistic texts.

\begin{figure}[h]
  \centering
  \includegraphics[width=0.5\textwidth]{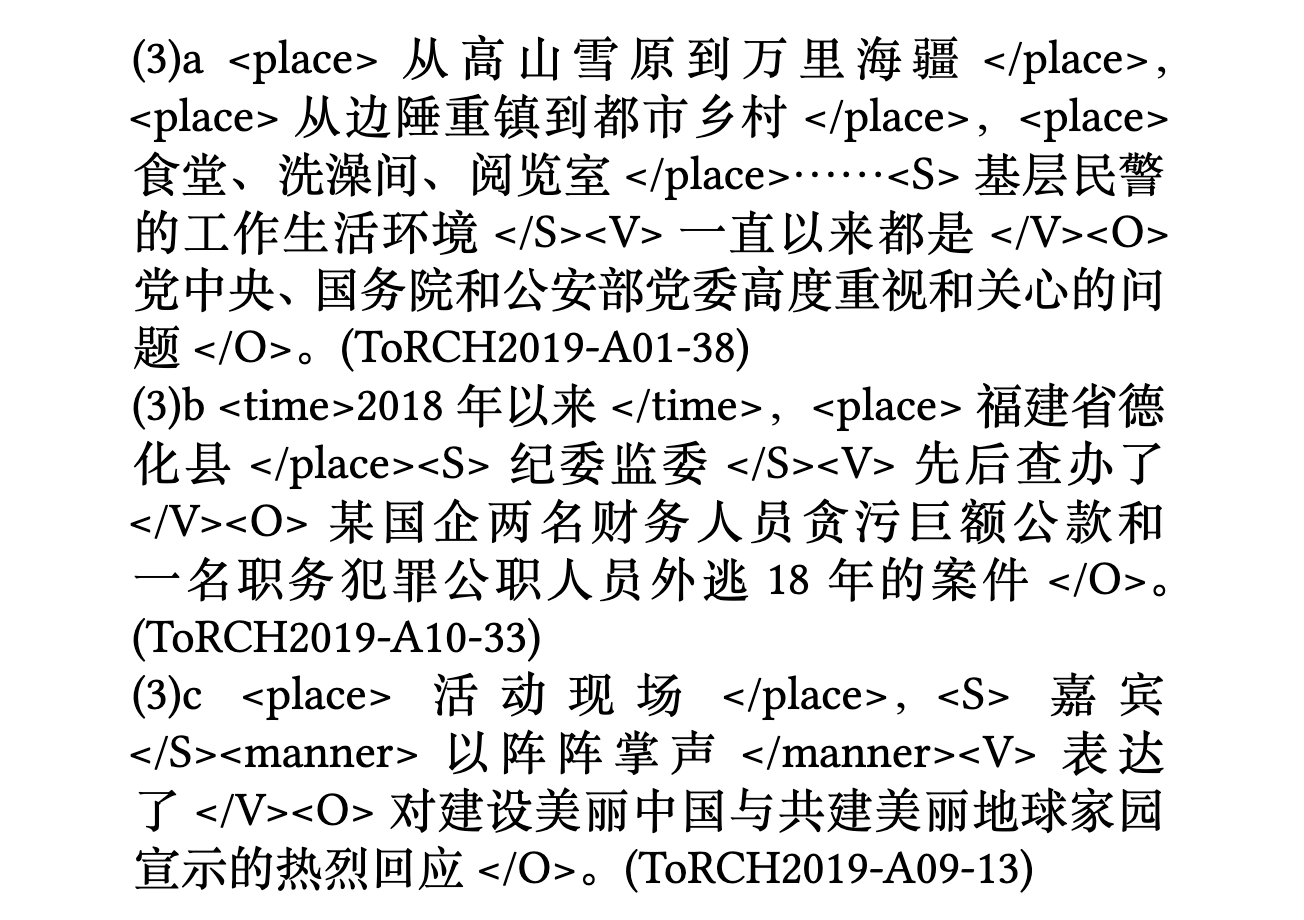}
  \label{fig:quote3}
\end{figure}

In English news, the distribution of functional chunks highlights their role in providing supplementary information through post-positioning. The <S><V><O><time> pattern exemplifies the post-positioning tendency of functional chunks. For instance, in example (4)a, the SVO structure first delivers the key information directly, followed by the gradual introduction of the <time> chunk, which adds specific background details. This arrangement enables readers to quickly grasp the core message, while the temporal background unfolds at the sentence’s end, avoiding overly lengthy initial modifiers that might disrupt the main structure.

Similarly, the <S><V><O><place> pattern, as seen in example (4)b, uses the SVO structure to present the event’s core content swiftly. The <place> chunk is then positioned at the end, progressively supplementing the spatial details of the event. This structure ensures a concise and straightforward narrative of the main event, while the spatial context unfolds gradually at the sentence’s end, clarifying the setting without interfering with the core information.

The <S><V><O><time><cause> pattern illustrates how functional chunks contribute to logical relationships. For example, in (4)c, <time> and <cause> chunks are sequentially appended at the sentence’s end, enriching the background and logical connections step by step. Although this progressive post-positioning may not strictly adhere to chronological logic, it maintains the simplicity of the sentence’s core structure while incrementally adding depth to the context and reasoning, making the information easier to process.

These examples clearly demonstrate that English news employs functional chunk post-positioning to achieve an effective strategy of ``quickly presenting SVO information, followed by gradual elaboration of the background.'' The pre-positioning of SVO and the post-positioning of functional chunks complement each other, ensuring a narrative that is both concise and efficient in logical progression while providing ample space for detailed background information. This feature not only enhances the readability of sentences but also strengthens the hierarchical organization of information.

\vspace{0.8cm}

\begin{quote}
(4)a <S>Neal</S><V>will assume</V><O>the helm of the nation's immigration court system</O><time>at a time when the Biden administration is wrestling with increasing arrivals at the US-Mexico border and seeking to reform a system that's been plagued with issues</time>. (CROWN2021-A03AA)

(4)b <S>Cuyjet (pronounced SOO-zhay)</S><V>also has</V><O>a video piece, ``For All Your Life Studies,''</O><place>in an exhibition called ``In Practice: You may go, but this will bring you back,'' at SculptureCenter in Long Island City</place>. (CROWN2021-A43AD)

(4)c <S>The White House</S><V>announced</V><O>a series of sanctions</O><time>on Friday night</time> \quad <cause>for its forced landing of a Ryanair commercial flight and the subsequent removal and arrest of opposition journalist Roman Protasevich</cause>. (CROWN2021-A04AE)
\end{quote}

\vspace{0.8cm}

\subsection{Sequential Patterns and Distribution of Multiple Functional Chunks}
We first analyzed the combination patterns of multiple functional chunks and compiled high-frequency co-occurrences (see Appendix Tables \ref{tab:table19} and \ref{tab:table21}). Among the top 10 combinations in both languages (see Tables \ref{tab:table12} and \ref{tab:table13}), <time><place> is the most frequent pair, indicating a certain degree of similarity in how the two languages arrange functional chunks. Overall, both Chinese and English exhibit significant flexibility in the arrangement of functional chunks.

In Chinese, combinations such as <place><manner> and <place><time> occur frequently, whereas English high-frequency combinations are more focused on causal and purposive functional chunks, such as <time><cause> and <effect><time>. However, these differences do not show significant interlinguistic variation, which may be attributable to the scope and size of the corpora. It is worth noting that combinations involving three functional chunks also appear in the data, but they generally lack statistical significance across languages. This further underscores the flexibility of combination patterns, as they are not concentrated in specific complex configurations.

\begin{figure*}[htbp]
\centering
\includegraphics[width=0.9\textwidth]{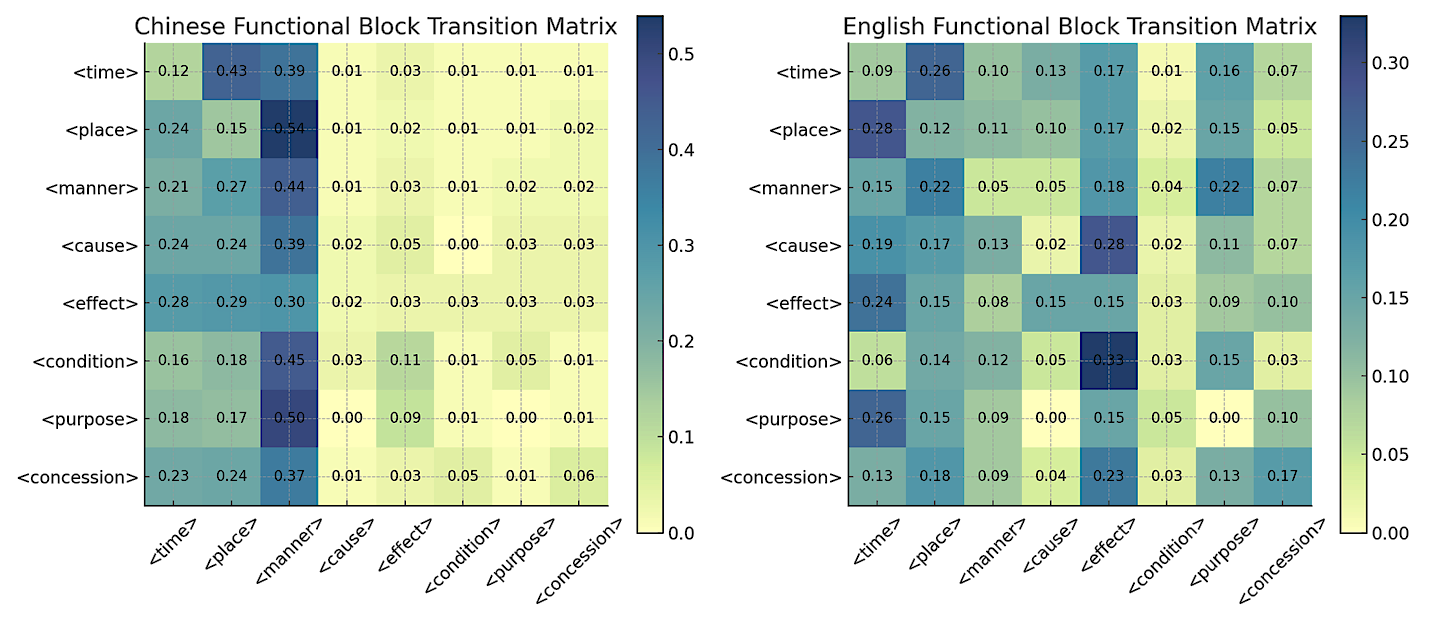}
\caption{Heatmaps of Functional Chunk Transition Matrix}
\label{fig:heatmap}
\end{figure*}

Inspired by Markov chains, we conducted an analysis of the conditional transition probabilities between functional chunks (see Appendix Tables \ref{tab:table23} - \ref{tab:table26}). Based on this analysis, we constructed transition matrix heatmaps (see Figure \ref{fig:heatmap}).

\begin{table}[htbp]
\centering
\caption{Top 10 Multi-Functional Chunk Combinations in Chinese}
\label{tab:table12}
\begin{tabular}{lll}
\toprule
Rank & Combination & Frequency \\
\midrule
1 & <time><place> & 415 \\
2 & <place><manner> & 345 \\
3 & <place><time> & 200 \\
4 & <manner><place> & 183 \\
5 & <time><manner> & 155 \\
6 & <manner><time> & 129 \\
7 & <time><place><manner> & 55 \\
8 & <effect><time> & 35 \\
9 & <place><manner><place> & 30 \\
10 & <manner><place><manner> & 28 \\
\bottomrule
\end{tabular}
\end{table}

\begin{table}[htbp]
\centering
\caption{Top 10 Multi-Functional Chunk Combinations in English}
\label{tab:table13}
\begin{tabular}{lll}
\toprule
Rank & Combination & Frequency \\
\midrule
1 & <time><place> & 152 \\
2 & <place><time> & 135 \\
3 & <time><cause> & 88 \\
4 & <effect><time> & 82 \\
5 & <time><purpose> & 73 \\
6 & <effect><cause> & 69 \\
7 & <purpose><time> & 64 \\
8 & <time><effect> & 63 \\
9 & <place><purpose> & 58 \\
10 & <place><effect> & 55 \\
\bottomrule
\end{tabular}
\end{table}

By comparing the top 10 transition probabilities (see Tables 9 and 10), we found that in Chinese, the highest transition probability is from <place> to <manner>, reaching 0.55. Additionally, <purpose> to <manner> and <condition> to <manner> are also among the top three transitions. In contrast, English shows higher transition probabilities from <condition> to <effect> (0.33) and <cause> to <effect> (0.28).

The heatmap further reveals that in Chinese, functional chunks such as <time>, <place>, and <manner> dominate transitions, highlighting Chinese's preference for using these chunks to organize information and guide narrative logic. This reflects the distinctive emphasis on scene-building and contextual layering in Chinese journalistic discourse. On the other hand, English exhibits a more evenly distributed transition pattern, with <time> and <effect> showing slightly higher prominence, indicating English's focus on logical connections to drive linear information progression.

Overall, both languages exhibit certain tendencies in the arrangement and distribution of multiple functional chunks. However, these preferences are not sufficiently consistent to form universally applicable conclusions due to the overall flexibility of usage. Combining the analysis of functional chunk combination patterns and transition matrices, we infer that this flexibility likely reflects the absence of rigid constraints in either language on functional chunk pairing, allowing them to be freely combined based on contextual needs.

\begin{table}[htbp]
\centering
\caption{Top 10 Functional Chunk Transitions in Chinese}
\label{tab:table14}
\begin{tabular}{lll}
\toprule
Rank & Transition & Probability \\
\midrule
1 & <place> to <manner> & 0.54 \\
2 & <purpose> to <manner> & 0.50 \\
3 & <condition> to <manner> & 0.45 \\
4 & <manner> to <manner> & 0.44 \\
5 & <time> to <place> & 0.43 \\
6 & <time> to <manner> & 0.39 \\
7 & <cause> to <manner> & 0.39 \\
8 & <concession> to <manner> & 0.37 \\
9 & <effect> to <manner> & 0.30 \\
10 & <effect> to <place> & 0.29 \\
\bottomrule
\end{tabular}
\end{table}

\begin{table}[htbp]
\centering
\caption{Top 10 Functional Chunk Transitions in English}
\label{tab:table15}
\begin{tabular}{lll}
\toprule
Rank & Transition & Probability \\
\midrule
1 & <condition> to <effect> & 0.33 \\
2 & <cause> to <effect> & 0.28 \\
3 & <place> to <time> & 0.28 \\
4 & <purpose> to <time> & 0.26 \\
5 & <time> to <place> & 0.26 \\
6 & <effect> to <time> & 0.24 \\
7 & <concession> to <effect> & 0.23 \\
8 & <manner> to <place> & 0.22 \\
9 & <manner> to <purpose> & 0.22 \\
10 & <cause> to <time> & 0.19 \\
\bottomrule
\end{tabular}
\end{table}
\section{Discussion}
\subsection{Functional Correspondence and Deep-Level Drivers of Information Organization}
The statistical results reveal significant preferences in the distribution of adverbial functional chunks within English and Chinese journalistic texts. The main differences lie in the sequence of presenting core events and background information when achieving the same informational functions. 

English typically adopts a “core information first” linear narrative mode, immediately establishing the skeletal structure of events and then elaborating on temporal, spatial, and other background details. This aligns with \citet{dryer1986}'s topic prominence hypothesis, which suggests that English prioritizes placing grammatical cores in prominent positions to enhance explicit information expression. Such a mode sharpens the narrative focus and logical coherence, making core content more identifiable and memorable.

In contrast, Chinese favors a “background first” holistic narrative approach. Common expressions include providing contextual information such as time, place, or manner to frame the narrative, while core events are often positioned at the end as the “highlight,” emphasizing progressive information delivery and contextual coherence. This characteristic validates \citet{rongjing2000}'s background–focus structure hypothesis, which posits that Chinese tends to support core events with contextual groundwork and guides readers through the narrative logic via gradual information progression. 

From an empirical perspective, these constituent order preferences support \citet{wangkangmao1981} descriptions of English and Chinese word order characteristics: English emphasizes informational concentration, while Chinese highlights semantic holism.

The difference between “core information first” and “background first” reflects not only surface syntactic features of the two languages but also deeper cognitive, semantic, and cultural mechanisms, offering multidimensional theoretical significance and practical value.

From a syntactic perspective \cite{ji2015}, English tends toward hypotaxis, with syntactic relations marked by strict formal rules that ensure linear progression and logical coherence. In actual corpora, background information such as <time> and <place> often appears post-verbally, making the core events within the SVO framework more prominent on the logical mainline. 

Conversely, Chinese relies on parataxis, with syntactic relations expressed more through semantic associations and contextual inference. In corpora, functional chunks like <time> and <place> frequently occur at the beginning of sentences, constructing a complete narrative background before progressively introducing the event core. This underscores Chinese’s reliance on holistic semantics and contextual connectivity.

Cognitive and cultural traditions further deepen the divergent preferences in English and Chinese word order. English cognitive patterns emphasize “objectivity and individuation,” centering on clear subjects and predicates to distinguish events from background information. This reflects the English cultural tradition of individual-centric and logical reasoning, which prioritizes the action logic of core events and postpones background details for gradual embellishment. 

In contrast, Chinese cognitive patterns emphasize holistic harmony, stressing the interaction between environment and actions \cite{jiadelin1990,zengjinghan2015}. Culturally, Chinese places greater emphasis on the interplay between context and behavior, favoring narratives that begin with background information to showcase the interaction between events and settings. For instance, reports on major events often begin with expansive temporal and spatial narratives before introducing specific actions or outcomes \cite{scollon2000}. The frequent fronting of background components establishes a semantic network and contextual framework that enriches the presentation of core events.

It is important to note that the observed characteristics are significantly influenced by the journalistic register. Future studies could expand the scope of the corpus and explore other genres to investigate the cross-register consistency and variability of English and Chinese constituent order patterns. Such efforts could provide deeper theoretical support for comparative linguistics and cross-cultural communication research.

\subsection{Exploring the Influence of Semantics on Word Order}
The conclusions discussed earlier are primarily based on the statistical analysis of functional chunk labels, largely decoupling word order patterns from specific sentence-level and functional chunk semantics. While the distribution of functional chunks reveals word order preferences in both languages, and we hypothesize that differences in functional chunk distribution are minimally tied to semantics, reflecting more of a cognitive construction, the potential influence of semantic differences on word order cannot be entirely dismissed.

Historically, large-scale semantic analysis of corpora has been challenging due to the time and labor required. However, modern large language models, trained on massive corpora, encode deep semantic representations within their parameters, enabling the embedding of corpora as high-dimensional vectors and facilitating efficient semantic analysis. To this end, we employed the MiniCPM-Embedding\footnote{\url{https://huggingface.co/openbmb/MiniCPM-Embedding}. MiniCPM-Embedding is a bilingual text embedding model jointly developed by ModelBest, TsinghuaNLP, and Information Retrieval Group at Northeastern University. All corpus embeddings were done on four RTX 4090D 24G GPUs.} model to selectively embed the corpora and analyze the resulting high-dimensional feature vectors.

\begin{figure*}[htbp]
\centering
\begin{subfigure}[b]{0.48\linewidth}
\includegraphics[width=\linewidth]{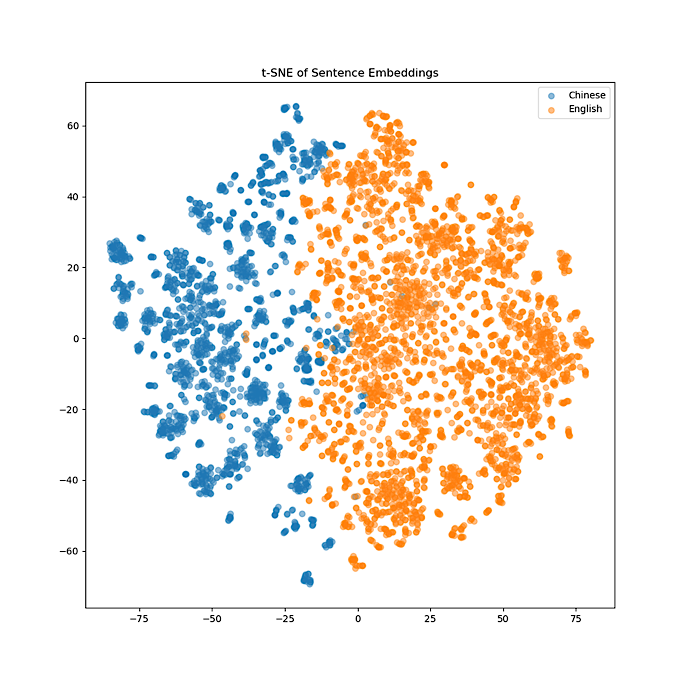}
\caption{Semantic Feature Distribution of two Corpora}
\label{fig:semantic_subcorpora}
\end{subfigure}
\hfill
\begin{subfigure}[b]{0.48\linewidth}
\includegraphics[width=\linewidth]{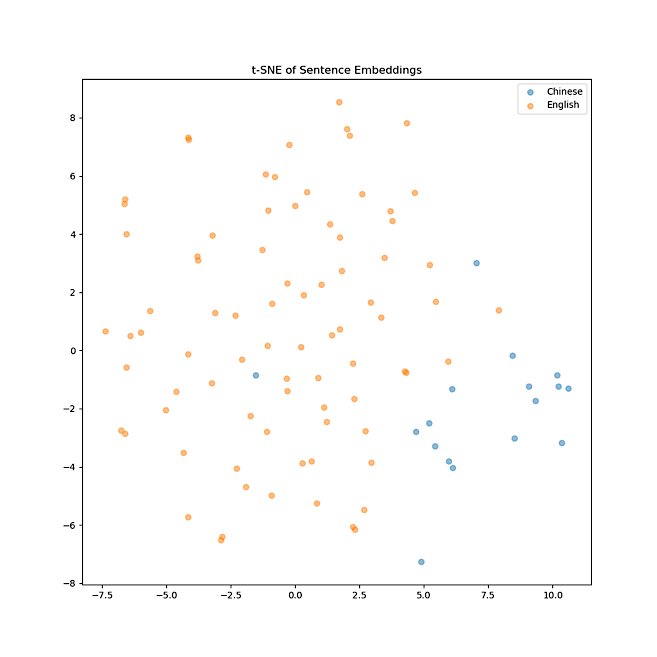}
\caption{Semantic Feature Distribution of Two Subsets}
\label{fig:semantic_subsets}
\end{subfigure}
\caption{Semantic Feature Distribution Analysis}
\label{fig:semantic_analysis}
\end{figure*}

First, we embedded both news subcorpora as well as subsets representing two typical English and Chinese word order structures (<time><S><V><O> and <S><V><O><time>). After deriving the central vectors, we visualized them using t-SNE dimensionality reduction and calculated cosine similarity\footnote{t-SNE (t-distributed Stochastic Neighbor Embedding) is a nonlinear dimensionality reduction technique for high-dimensional data visualization. It preserves the local neighborhood relationships between high-dimensional data points, effectively showcasing the local structural features of the data. Cosine similarity, on the other hand, is calculated based on the directional relationships of global feature vectors. It reveals the overall similarity of the data while being less sensitive to localized differences.}. The cosine similarity of the complete English and Chinese news corpora was 74.07\%, while the similarity for subsets related to the functional chunk was 62.04\%, indicating some semantic differences between the English and Chinese news corpora, though they are generally quite similar overall. However, as shown in Figure \ref{fig:semantic_analysis}, noticeable local semantic differences exist within the corpora.

Combining these findings with the earlier conclusions on word order preferences, we infer that while the overall semantic differences in the corpora are minimal, localized differences do exist. Thus, we can only conclude that semantics has a relatively minor influence on the differences in word order between English and Chinese journalistic texts, but the role of semantic differences, particularly local ones, should not be overlooked. This suggests that the distribution of semantic features has a potential impact on word order choices, with a limited overall influence but possible significance in specific contexts.

\begin{figure*}[htbp]
\centering
\caption{Semantic Features Distributions of Functional Chunks}
\begin{subfigure}[b]{0.24\linewidth}
\includegraphics[width=\linewidth]{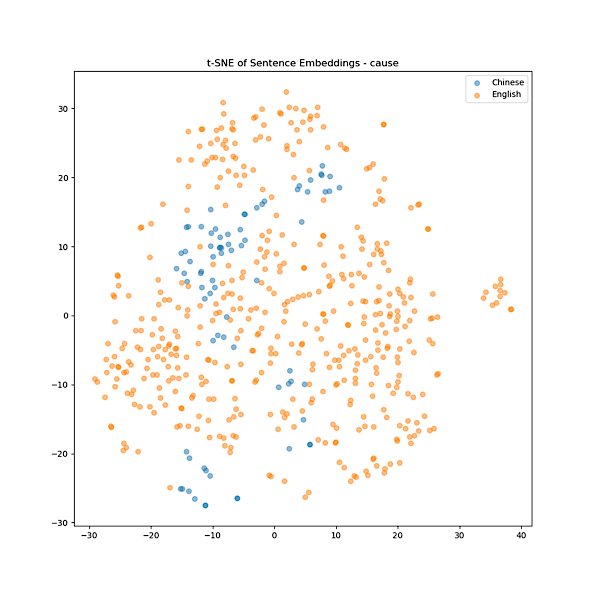}
\label{fig:sub23}
\end{subfigure}
\hfill
\begin{subfigure}[b]{0.24\linewidth}
\includegraphics[width=\linewidth]{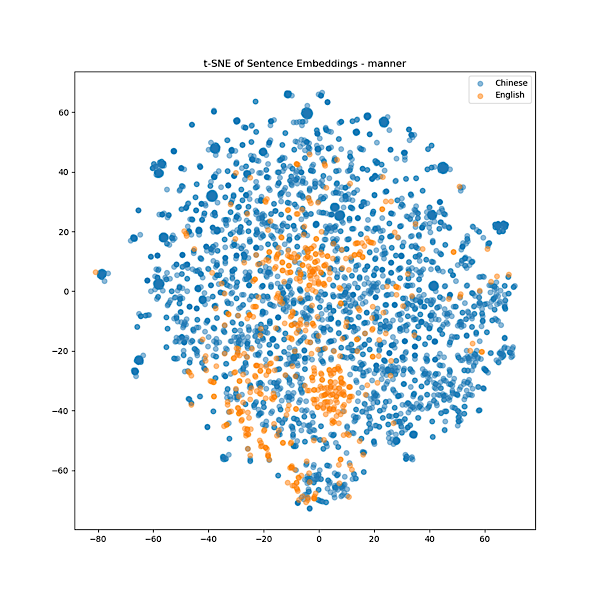}
\label{fig:sub24}
\end{subfigure}
\hfill
\begin{subfigure}[b]{0.24\linewidth}
\includegraphics[width=\linewidth]{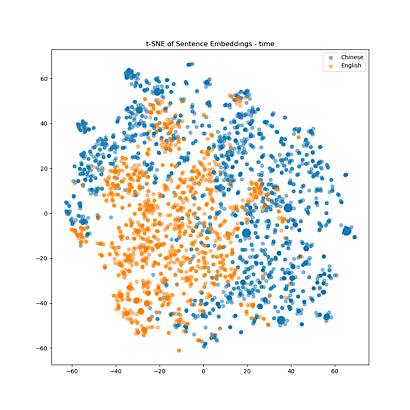}
\label{fig:sub25}
\end{subfigure}
\hfill
\begin{subfigure}[b]{0.24\linewidth}
\includegraphics[width=\linewidth]{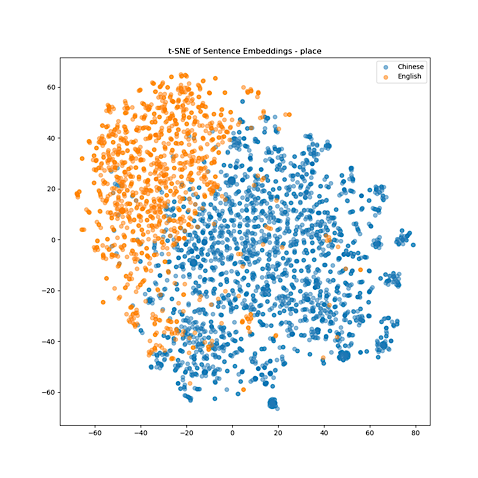}
\label{fig:sub26}
\end{subfigure}


\begin{subfigure}[b]{0.24\linewidth}
\includegraphics[width=\linewidth]{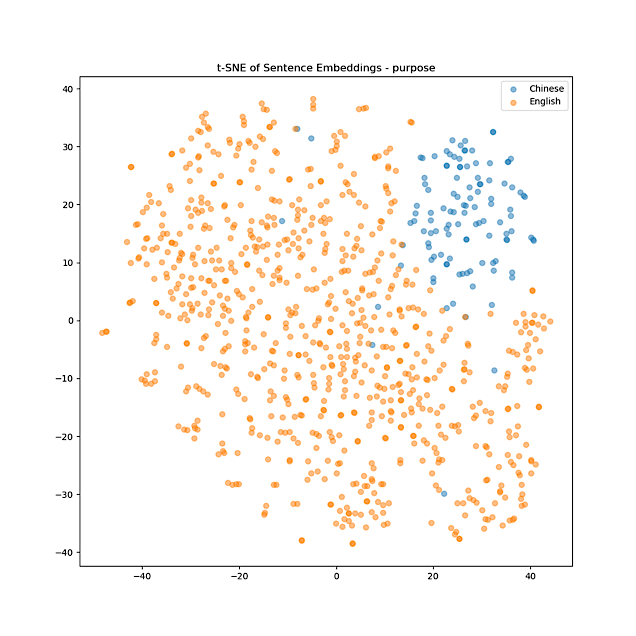}
\label{fig:sub27}
\end{subfigure}
\hfill
\begin{subfigure}[b]{0.24\linewidth}
\includegraphics[width=\linewidth]{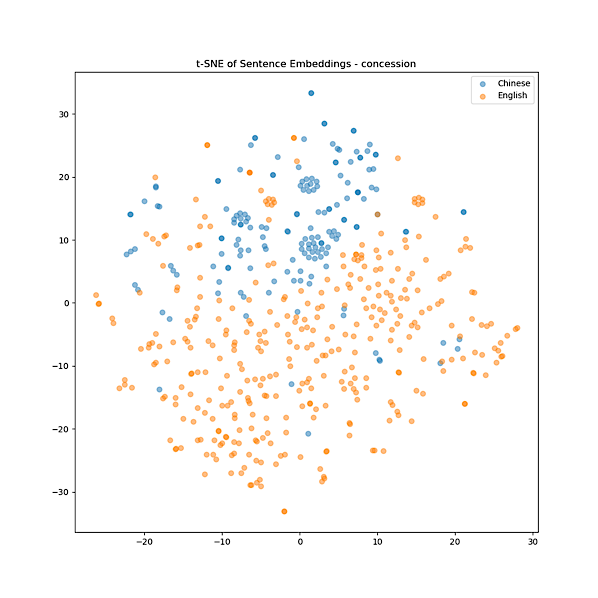}
\label{fig:sub28}
\end{subfigure}
\hfill
\begin{subfigure}[b]{0.24\linewidth}
\includegraphics[width=\linewidth]{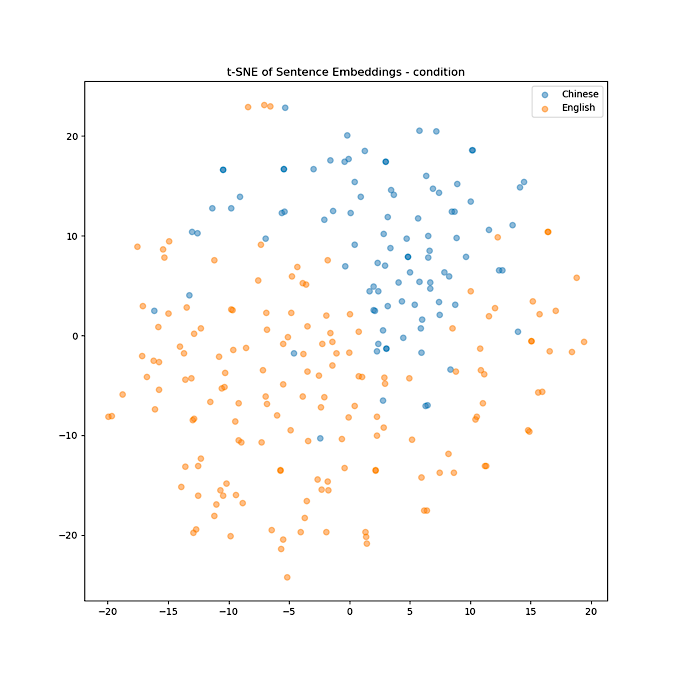}
\label{fig:sub29}
\end{subfigure}
\hfill
\begin{subfigure}[b]{0.24\linewidth}
\includegraphics[width=\linewidth]{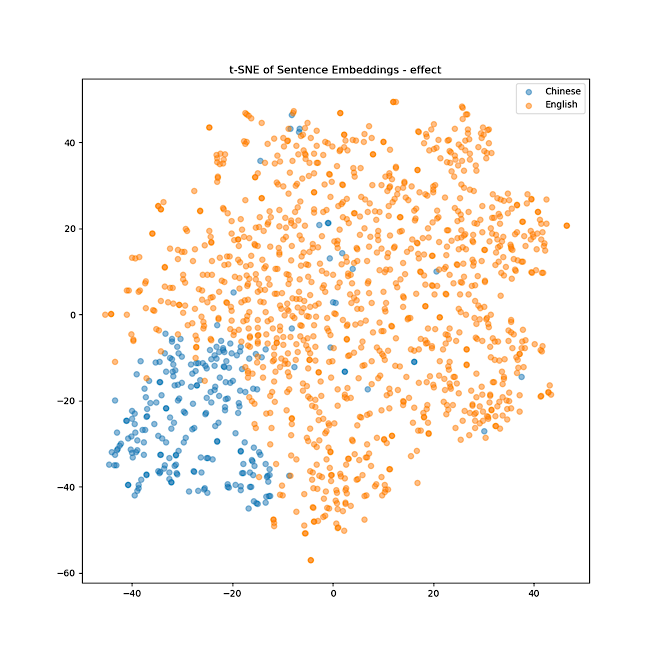}
\label{fig:sub30}
\end{subfigure}

\label{fig:combined_images_23_30}
\end{figure*}

We further applied the same methodology to compute the semantic similarity of the specific corpus content corresponding to the eight types of functional chunks. The overall semantic similarity for each functional chunk exceeded or approached 85\% (see Table \ref{tab:table16}). Functional chunks such as <time>, <manner>, <cause>, and <concession> exhibited relatively high overlap in their local distributions, while other categories showed slight distributional differences (see Figure \ref{fig:combined_images_23_30}).

This indicates that the semantic differences between functional chunks themselves are minimal, suggesting that the contribution of chunk-level semantics to the preferences for functional chunk distribution is relatively small. However, we cannot entirely dissociate semantics from word order. The overall semantic similarity of functional chunks may obscure local differences, and these local variations could potentially drive specific word order choices.

While embeddings effectively capture the semantic features of corpora, they also have limitations. Their ability to represent deep micro-level semantic associations is restricted, and they may overlook dynamic pragmatic factors in high-dimensional spaces. Through variable control, we can conclude that ``semantics has a limited but non-negligible impact on word order differences,'' but we have not further refined the specific mechanisms by which semantics influences word order.

Future research could explore a broader range of corpora and functional chunks while incorporating dynamic semantic modeling to investigate the complex interaction between semantics and word order. Such studies would provide a more detailed theoretical framework and empirical support for research on English and Chinese word order.

\begin{table}[htbp]
\centering
\caption{Sematic Similarities of Functional Chunks}
\label{tab:table16}
\begin{tabular}{ll}
\toprule
FC & Cosine Similarity (\%) \\
\midrule
<time> & 91.19 \\
<place> & 88.47 \\
<manner> & 86.61 \\
<cause> & 84.92 \\
<effect> & 87.47 \\
<condition> & 88.01 \\
<purpose> & 85.29 \\
<concession> & 86.63 \\
\bottomrule
\end{tabular}
\end{table}

\subsection{Mechanisms of Flexibility within Fixed Preferences}
The distribution of constituent order in English and Chinese journalistic texts demonstrates significant patterns of regularity. However, we argue that these patterns are by no means rigid but exhibit a degree of flexibility. This flexibility is influenced by factors such as information structure, pragmatic needs, and specific contexts, reflecting the dual characteristics of preference and adaptability in constituent order for both languages. This observation provides new empirical support for the scholarly debate on word order.

For example, in English corpora, when certain background elements carry important discourse functions, their position may shift from post-positioning to pre-positioning. For instance:
\vspace{0.8cm}

\begin{quote}
The public defense of the court as a nonpartisan institution comes at a fraught time for the justices and their credibility. The Court's approval rating has dipped below 50\% for the first time since 2017 and down 9-points from a decade high just last year, according to Gallup. <time>This month</time>, <S>the court</S> <V>became embroiled</V> <O>in a dramatic and highly divisive debate</O> <cause>over abortion in Texas</cause> (CROWN2021-A12AB)
\end{quote}
\vspace{0.8cm}

In this sentence, the earlier context discussed the Supreme Court's ``precarious'' situation since 2017 and the public's skepticism about its impartiality. The concluding sentence naturally introduces ``this month'' as a specific case reflecting this trust crisis, transitioning from an abstract temporal context to a concrete fact. The pre-positioning of the temporal element highlights its bridging role in the discourse, demonstrating English's flexibility in adapting word order to pragmatic demands.

\vspace{0.1cm}

Similarly, Chinese news texts also exhibit flexible word order arrangements. For instance:

\begin{figure}[h]
  \centering
  \includegraphics[width=0.48\textwidth]{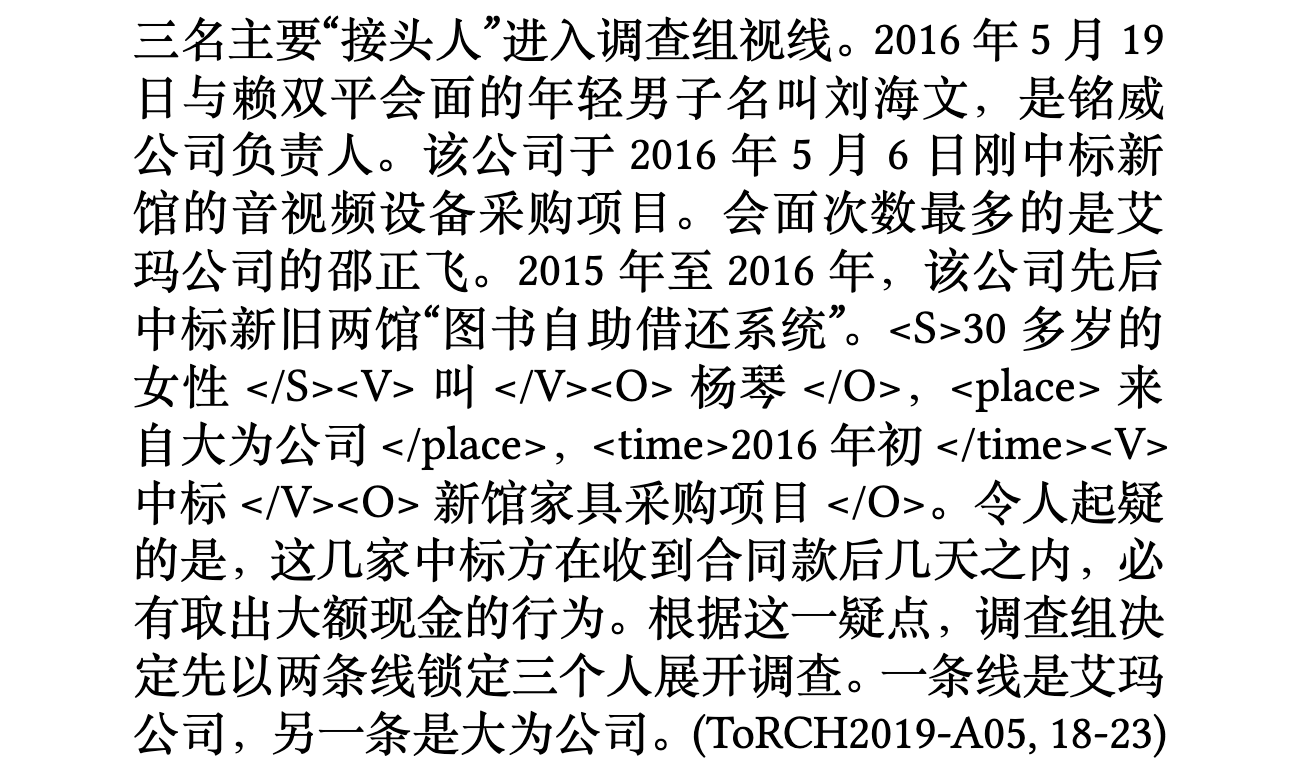}
  \label{fig:quote4}
\end{figure}

In this example, the core information revolves around the identity of the woman and her bidding actions. Placing the SVO structure at the forefront aligns with the demand for focus-first expression, while <time> and <place> are placed at the sentence’s end to create a causal relationship. From a discourse perspective, the narrative adopts a general-specific-general structure, with the roles and actions of three ``contact persons'' forming a parallel relationship. Adjusting <place> and <time> to follow the core SVO structure aligns with readers’ logical information reception processes, enhancing narrative layering. Although this distribution deviates from the general ``background first'' preference of Chinese, it adapts efficiently to highlight focus and strengthen logic.

The scholarly debate over whether word order follows fixed rules often centers on the relationship between syntax and information structure. On one hand, traditional research suggests that constituent order is typologically influenced, with syntax significantly constraining information structure. On the other hand, recent studies emphasize the pragmatic drivers of word order, arguing that choices are more dependent on information focus, topic progression, and contextual fit.

Through comparative analysis of English and Chinese journalistic corpora, this study finds that constituent order exhibits stable distributional regularities but also demonstrates flexibility under pragmatic demands. This flexibility reflects the adaptability of language to cognitive and informational organization needs. For example, while English adheres to the dependency locality principle favoring linear progression, certain contexts may pre-position background information to emphasize topics or reduce cognitive load. Similarly, while Chinese’s ``topic chain'' structure prioritizes holistic elaboration, it may post-position background information when focus needs to be highlighted.

These phenomena collectively illustrate that word order balances both regularity and flexibility. It exhibits systematic preferences while retaining the ability to adjust dynamically. This dual characteristic provides a balanced explanation for the debate over whether word order is ``fixed.'' Word order choices are largely influenced by pragmatic needs, information focus, and contextual logic. This not only broadens our understanding of English and Chinese word order features but also offers valuable directions for further research into cross-register word order patterns.

\section{Conclusion}
This study, based on a large-scale English-Chinese journalistic corpus and utilizing automatic annotation by large language models, explored the constituent order preferences and arrangement patterns in the two languages from the perspective of functional chunks. The findings are as follows:

(1) Functional chunks such as time and place exhibit significant word order preferences in English and Chinese journalistic texts. English tends to position these chunks at the end of sentences, emphasizing the informational centrality of core events. In contrast, Chinese demonstrates a stronger fronting tendency, highlighting the importance of background information in setting the narrative stage.

(2) The combination patterns of the SVO structure with functional chunks reflect differences in word order preferences between English and Chinese journalism. These differences vary in intensity: Chinese strongly favors fronting, while English exhibits a more moderate inclination toward post-positioning, reflecting divergent focuses in information organization.

(3) When multiple functional chunks co-occur, both English and Chinese journalistic texts display high flexibility. Word order adjustments are dynamically influenced by information focus, pragmatic needs, and discourse functions.

These findings partially reveal the differences in word order preferences and information organization strategies in English and Chinese journalistic texts. They validate typological hypotheses on word order distribution from existing theories and provide new empirical support for the pragmatic adaptation mechanisms of word order.

\section*{Limitations}

We acknowledges several methodological limitations that warrant consideration for future research:

\textbf{Annotation Consistency and Human-Machine Alignment.} The reliability of LLM annotations remains challenging to achieve consistently across the corpus. Additionally, the lack of systematic alignment between human and machine annotation standards may introduce discrepancies.

\textbf{Taxonomic Completeness.} The current annotation scheme does not fully satisfy the MECE (Mutually Exclusive, Collectively Exhaustive) principle, resulting in potential gaps and overlaps in functional chunk categorization. More rigorous definitions and systematic rules are needed.

\textbf{Prompt-Induced Bias.} The design of annotation prompts may inadvertently introduce researcher bias into the model's output, potentially influencing results in ways that do not reflect objective linguistic phenomena.

\textbf{Semantic Understanding Limitations.} Current LLMs demonstrate fundamental differences from human linguistic comprehension, particularly in processing implicit meanings, metaphorical expressions, and culturally embedded semantic content. Surface-level embeddings may not adequately capture the deeper semantic nuances present in authentic discourse.

\textbf{Scope and Scale Constraints.} As an undergraduate research project conducted within the constraints of coursework requirements, this study operates under significant temporal and resource limitations that necessarily restrict the depth and breadth of the investigation.

\section*{Acknowledgments}

This work has been independently conducted and submitted to (1) \textit{Corpus Linguistics, English-Chinese Contrastive Linguistics,} and \textit{Natural Language Processing Fundamentals} at School of Humanities, BUPT in 2024 Fall; (2) \textit{Linguistic Data and Python Applications} with edits in 2025 Spring; and (3) \textit{The First Undergraduate Academic Innovation Forum of Beijing Foreign Studies University} on April 23, 2025.

\bibliographystyle{ACM-Reference-Format}
\bibliography{sample-base}

\clearpage

\appendix

\begin{table}[h]
\centering
\caption{Top 20 Functional Chunk–SVO Patterns in Chinese}
\label{tab:table17}
\begin{tabular}{lll}
\toprule
Rank & Pattern & Frequency \\
\midrule
1 & <time><S><V><O> & 17 \\
2 & <time><V><O> & 15 \\
3 & <place><S><V><O> & 13 \\
4 & <manner><V><O> & 13 \\
5 & <place><V><O> & 12 \\
6 & <S><V><O><manner><V><O> & 7 \\
7 & <S><V><O><manner> & 7 \\
8 & <place><S><V> & 7 \\
9 & <effect><V><O> & 5 \\
10 & <V><O><V><O><manner> & 4 \\
11 & <place><S><manner><V><O> & 4 \\
12 & <time><S><V> & 4 \\
13 & <S><place><V><O> & 4 \\
14 & <time><place><S><V><O> & 4 \\
15 & <S><V><O><manner><O> & 3 \\
16 & <time><S><V><O><V><O> & 3 \\
17 & <S><V><O><place> & 3 \\
18 & <time><place><V><O> & 3 \\
19 & <place><manner><V><O> & 3 \\
20 & <time><S><V><place> & 3 \\
\bottomrule
\end{tabular}
\end{table}

\begin{table}[h]
\centering
\caption{Top 20 Functional Chunk–SVO Patterns in English}
\label{tab:table18}
\begin{tabular}{lll}
\toprule
Rank & Pattern & Frequency \\
\midrule
1 & <S><V><effect> & 127 \\
2 & <S><V><O><time> & 81 \\
3 & <S><V><O><effect> & 70 \\
4 & <S><V><O><place> & 61 \\
5 & <S><V><O><purpose> & 59 \\
6 & <S><V><place> & 51 \\
7 & <S><V><time> & 45 \\
8 & <S><V><purpose> & 43 \\
9 & <S><V><O><concession> & 33 \\
10 & <S><V><O><manner> & 32 \\
11 & <S><V><O><cause> & 23 \\
12 & <S><V><effect><time> & 21 \\
13 & <S><V><effect><cause> & 20 \\
14 & <S><V><time><place> & 20 \\
15 & <S><V><cause> & 18 \\
16 & <S><V><O><time><cause> & 17 \\
17 & <S><V><time><cause> & 15 \\
18 & <S><V><effect><concession> & 15 \\
19 & <S><V><time><effect> & 14 \\
20 & <S><V><O><place><time> & 14 \\
\bottomrule
\end{tabular}
\end{table}

\clearpage

\begin{table}[htbp]
\centering
\caption{Top 50 Multi-Functional Chunk Combinations in Chinese}
\label{tab:table19}
\begin{tabular}{lll}
\toprule
Rank & Combination & Frequency \\
\midrule
1 & <time><place> & 415 \\
2 & <place><manner> & 345 \\
3 & <place><time> & 200 \\
4 & <manner><place> & 183 \\
5 & <time><manner> & 155 \\
6 & <manner><time> & 129 \\
7 & <time><place><manner> & 55 \\
8 & <effect><time> & 35 \\
9 & <place><manner><place> & 30 \\
10 & <manner><place><manner> & 28 \\
11 & <manner><effect> & 25 \\
12 & <effect><manner> & 25 \\
13 & <effect><place> & 24 \\
14 & <manner><time><place> & 23 \\
15 & <concession><place> & 21 \\
16 & <manner><concession> & 20 \\
17 & <concession><manner> & 19 \\
18 & <purpose><manner> & 18 \\
19 & <place><time><place> & 18 \\
20 & <condition><manner> & 17 \\
21 & <place><manner><time> & 16 \\
22 & <place><time><manner> & 16 \\
23 & <time><place><time> & 13 \\
24 & <time><effect> & 12 \\
25 & <time><manner><place> & 11 \\
26 & <manner><condition> & 11 \\
27 & <manner><time><manner> & 11 \\
28 & <place><manner><place><manner> & 10 \\
29 & <manner><purpose> & 10 \\
30 & <concession><time> & 10 \\
31 & <time><condition> & 9 \\
32 & <time><cause> & 9 \\
33 & <cause><manner> & 8 \\
34 & <time><purpose> & 8 \\
35 & <purpose><time> & 8 \\
36 & <cause><place> & 7 \\
37 & <cause><time> & 7 \\
38 & <purpose><place> & 7 \\
39 & <place><cause> & 7 \\
40 & <purpose><effect> & 6 \\
41 & <effect><time><place> & 6 \\
42 & <manner><cause> & 6 \\
43 & <manner><place><manner><place> & 6 \\
44 & <manner><place><time> & 6 \\
45 & <condition><place> & 6 \\
46 & <time><manner><time> & 6 \\
47 & <place><purpose> & 6 \\
48 & <effect><place><manner> & 6 \\
49 & <condition><effect> & 5 \\
50 & <time><place><manner><place> & 5 \\
\bottomrule
\end{tabular}
\end{table}

\begin{table}[htbp]
\centering
\caption{Top 50 Multi-Functional Chunk Combinations in English}
\label{tab:table21}
\begin{tabular}{lll}
\toprule
Rank & Combination & Frequency \\
\midrule
1 & <time><place> & 152 \\
2 & <place><time> & 135 \\
3 & <time><cause> & 88 \\
4 & <effect><time> & 82 \\
5 & <time><purpose> & 73 \\
6 & <effect><cause> & 69 \\
7 & <purpose><time> & 64 \\
8 & <time><effect> & 63 \\
9 & <place><purpose> & 58 \\
10 & <place><effect> & 55 \\
11 & <manner><place> & 53 \\
12 & <place><manner> & 53 \\
13 & <manner><purpose> & 51 \\
14 & <place><cause> & 48 \\
15 & <time><concession> & 45 \\
16 & <time><manner> & 43 \\
17 & <effect><concession> & 43 \\
18 & <effect><place> & 40 \\
19 & <purpose><place> & 37 \\
20 & <purpose><concession> & 36 \\
21 & <manner><time> & 35 \\
22 & <manner><effect> & 31 \\
23 & <place><concession> & 31 \\
24 & <purpose><cause> & 27 \\
25 & <concession><cause> & 25 \\
26 & <effect><purpose> & 23 \\
27 & <manner><concession> & 23 \\
28 & <purpose><manner> & 23 \\
29 & <cause><effect> & 22 \\
30 & <purpose><effect> & 21 \\
31 & <effect><manner> & 19 \\
32 & <cause><time> & 16 \\
33 & <purpose><condition> & 14 \\
34 & <concession><effect> & 14 \\
35 & <condition><effect> & 14 \\
36 & <concession><place> & 12 \\
37 & <cause><concession> & 12 \\
38 & <time><condition> & 11 \\
39 & <cause><purpose> & 11 \\
40 & <time><place><effect> & 11 \\
41 & <manner><cause> & 11 \\
42 & <time><place><manner> & 10 \\
43 & <effect><condition> & 10 \\
44 & <cause><place> & 10 \\
45 & <concession><purpose> & 10 \\
46 & <place><time><cause> & 9 \\
47 & <cause><manner> & 9 \\
48 & <concession><time> & 9 \\
49 & <time><place><cause> & 9 \\
50 & <place><purpose><time> & 9 \\
\bottomrule
\end{tabular}
\end{table}

\clearpage

\begin{table}[htbp]
\centering
\caption{Transition Probabilities of Functional Chunks in Chinese}
\label{tab:table23}
\begin{tabular}{lll}
\toprule
Rank & Transition & Probability \\
\midrule
1 & <place> to <manner> & 0.55 \\
2 & <purpose> to <manner> & 0.50 \\
3 & <condition> to <manner> & 0.45 \\
4 & <manner> to <manner> & 0.44 \\
5 & <time> to <place> & 0.43 \\
6 & <time> to <manner> & 0.39 \\
7 & <cause> to <manner> & 0.39 \\
8 & <concession> to <manner> & 0.37 \\
9 & <effect> to <manner> & 0.30 \\
10 & <effect> to <place> & 0.29 \\
11 & <effect> to <time> & 0.28 \\
12 & <manner> to <place> & 0.27 \\
13 & <cause> to <time> & 0.24 \\
14 & <cause> to <place> & 0.24 \\
15 & <place> to <time> & 0.24 \\
16 & <concession> to <place> & 0.24 \\
17 & <concession> to <time> & 0.23 \\
18 & <manner> to <time> & 0.21 \\
19 & <condition> to <place> & 0.18 \\
20 & <purpose> to <time> & 0.18 \\
21 & <purpose> to <place> & 0.17 \\
22 & <condition> to <time> & 0.16 \\
23 & <place> to <place> & 0.15 \\
24 & <time> to <time> & 0.12 \\
25 & <condition> to <effect> & 0.11 \\
26 & <purpose> to <effect> & 0.09 \\
27 & <concession> to <concession> & 0.06 \\
28 & <condition> to <purpose> & 0.05 \\
29 & <purpose> to <condition> & 0.05 \\
30 & <concession> to <condition> & 0.05 \\
31 & <cause> to <effect> & 0.05 \\
32 & <concession> to <effect> & 0.04 \\
\bottomrule
\end{tabular}
\end{table}

\begin{table}[htbp]
\centering
\caption{(Continued) Transition Probabilities of Functional Chunks in Chinese}
\label{tab:table24}
\begin{tabular}{lll}
\toprule
Rank & Transition & Probability \\
\midrule
33 & <effect> to <purpose> & 0.03 \\
34 & <cause> to <purpose> & 0.03 \\
35 & <cause> to <concession> & 0.03 \\
36 & <manner> to <effect> & 0.03 \\
37 & <effect> to <effect> & 0.03 \\
38 & <effect> to <condition> & 0.03 \\
39 & <effect> to <concession> & 0.03 \\
40 & <condition> to <cause> & 0.03 \\
41 & <time> to <effect> & 0.03 \\
42 & <place> to <effect> & 0.02 \\
43 & <manner> to <concession> & 0.02 \\
44 & <effect> to <cause> & 0.02 \\
45 & <place> to <concession> & 0.02 \\
46 & <manner> to <purpose> & 0.02 \\
47 & <cause> to <cause> & 0.02 \\
48 & <concession> to <cause> & 0.01 \\
49 & <concession> to <purpose> & 0.01 \\
50 & <condition> to <condition> & 0.01 \\
51 & <condition> to <concession> & 0.01 \\
52 & <time> to <purpose> & 0.01 \\
53 & <purpose> to <concession> & 0.01 \\
54 & <manner> to <condition> & 0.01 \\
55 & <time> to <cause> & 0.01 \\
56 & <place> to <purpose> & 0.01 \\
57 & <time> to <concession> & 0.01 \\
58 & <place> to <cause> & 0.01 \\
59 & <place> to <condition> & 0.01 \\
60 & <time> to <condition> & 0.01 \\
61 & <manner> to <cause> & 0.01 \\
62 & <cause> to <condition> & 0.00 \\
63 & <purpose> to <cause> & 0.00 \\
64 & <purpose> to <purpose> & 0.00 \\
\bottomrule
\end{tabular}
\end{table}

\clearpage

\begin{table}[htbp]
\centering
\caption{Transition Probabilities of Functional Chunks in English}
\label{tab:table25}
\begin{tabular}{lll}
\toprule
Rank & Transition & Probability \\
\midrule
1 & <condition> to <effect> & 0.33 \\
2 & <cause> to <effect> & 0.28 \\
3 & <place> to <time> & 0.28 \\
4 & <purpose> to <time> & 0.26 \\
5 & <time> to <place> & 0.26 \\
6 & <effect> to <time> & 0.24 \\
7 & <concession> to <effect> & 0.23 \\
8 & <manner> to <place> & 0.22 \\
9 & <manner> to <purpose> & 0.22 \\
10 & <cause> to <time> & 0.19 \\
11 & <manner> to <effect> & 0.18 \\
12 & <concession> to <place> & 0.18 \\
13 & <time> to <effect> & 0.17 \\
14 & <concession> to <cause> & 0.17 \\
15 & <cause> to <place> & 0.17 \\
16 & <place> to <effect> & 0.17 \\
17 & <time> to <purpose> & 0.16 \\
18 & <condition> to <purpose> & 0.16 \\
19 & <purpose> to <place> & 0.15 \\
20 & <effect> to <cause> & 0.15 \\
21 & <manner> to <time> & 0.15 \\
22 & <effect> to <place> & 0.15 \\
23 & <effect> to <effect> & 0.15 \\
24 & <place> to <purpose> & 0.15 \\
25 & <condition> to <place> & 0.14 \\
26 & <purpose> to <effect> & 0.14 \\
27 & <concession> to <time> & 0.13 \\
28 & <concession> to <purpose> & 0.13 \\
29 & <time> to <cause> & 0.13 \\
30 & <cause> to <manner> & 0.13 \\
31 & <condition> to <manner> & 0.12 \\
32 & <place> to <place> & 0.12 \\
\bottomrule
\end{tabular}
\end{table}

\begin{table}[htbp]
\centering
\caption{(Continued) Transition Probabilities of Functional Chunks in English}
\label{tab:table26}
\begin{tabular}{lll}
\toprule
Rank & Transition & Probability \\
\midrule
33 & <place> to <manner> & 0.12 \\
34 & <cause> to <purpose> & 0.11 \\
35 & <purpose> to <concession> & 0.11 \\
36 & <time> to <manner> & 0.11 \\
37 & <effect> to <concession> & 0.10 \\
38 & <condition> to <concession> & 0.10 \\
39 & <purpose> to <purpose> & 0.10 \\
40 & <place> to <cause> & 0.10 \\
41 & <purpose> to <manner> & 0.10 \\
42 & <effect> to <purpose> & 0.09 \\
43 & <time> to <time> & 0.09 \\
44 & <purpose> to <cause> & 0.09 \\
45 & <concession> to <manner> & 0.09 \\
46 & <effect> to <manner> & 0.08 \\
47 & <cause> to <concession> & 0.07 \\
48 & <manner> to <concession> & 0.07 \\
49 & <time> to <concession> & 0.07 \\
50 & <condition> to <time> & 0.06 \\
51 & <manner> to <manner> & 0.06 \\
52 & <manner> to <cause> & 0.06 \\
53 & <condition> to <cause> & 0.05 \\
54 & <place> to <concession> & 0.05 \\
55 & <purpose> to <condition> & 0.05 \\
56 & <concession> to <condition> & 0.04 \\
57 & <manner> to <condition> & 0.04 \\
58 & <concession> to <concession> & 0.03 \\
59 & <condition> to <condition> & 0.03 \\
60 & <effect> to <condition> & 0.03 \\
61 & <cause> to <cause> & 0.02 \\
62 & <cause> to <condition> & 0.02 \\
63 & <place> to <condition> & 0.02 \\
64 & <time> to <condition> & 0.01 \\
\bottomrule
\end{tabular}
\end{table}

\end{document}